\PassOptionsToPackage{table,xcdraw}{xcolor}
\documentclass[11pt]{article}

\usepackage[preprint]{acl}

\usepackage{times}
\usepackage{latexsym}

\usepackage[T1]{fontenc}

\usepackage[utf8]{inputenc}

\usepackage{microtype}

\usepackage{inconsolata}

\usepackage{times}
\usepackage{soul}
\usepackage{url}
\usepackage{graphicx}
\usepackage{amsmath}
\usepackage{amsthm}
\usepackage{booktabs}
\usepackage{amsmath}
\usepackage{enumitem}
\usepackage{tabularx}
\usepackage{algorithm}
\usepackage{algorithmic}
\usepackage{caption}
\usepackage{cleveref}
\usepackage{titlesec}
\usepackage{lipsum}
\usepackage{tikz}
\usepackage{colortbl}
\usepackage{multirow}
\usepackage{array}
\usepackage{listings}
\usepackage{tcolorbox}

\usepackage[dvipsnames]{xcolor}

\newcommand{\hlA}[1]{\colorbox{RoyalBlue!15}{#1}}
\newcommand{\hlB}[1]{\colorbox{BrickRed!15}{#1}}
\newcommand{\hlC}[1]{\colorbox{ForestGreen!15}{#1}}
\definecolor{mygray}{gray}{.9}
\definecolor{codegreen}{rgb}{0,0.6,0}
\definecolor{codegray}{rgb}{0.5,0.5,0.5}
\definecolor{codepurple}{rgb}{0.58,0,0.82}
\definecolor{backcolour}{rgb}{0.95,0.95,0.92}

\definecolor{rulecolor}{rgb}{0.1,0.1,0.5}
\definecolor{commentcolor}{rgb}{0.5,0.5,0.5}
\definecolor{featurecolor}{rgb}{0.0,0.5,0.0}

\lstdefinestyle{pddlstyle}{
    backgroundcolor=\color{backcolour},   
    commentstyle=\color{codegreen},
    keywordstyle=\color{magenta},
    numberstyle=\tiny\color{codegray},
    stringstyle=\color{codepurple},
    basicstyle=\ttfamily\footnotesize,
    breakatwhitespace=false,         
    breaklines=true,                 
    captionpos=b,                    
    keepspaces=true,                 
    numbers=left,                    
    numbersep=5pt,                  
    showspaces=false,                
    showstringspaces=false,
    showtabs=false,                  
    tabsize=2,
    language=lisp
}

\crefname{section}{\S}{\S\S}
\Crefname{section}{\S}{\S\S}

\title{Mini-BEHAVIOR-Gran: Revealing U-Shaped Effects of Instruction Granularity on Language-Guided Embodied Agents}

\author{
 \textbf{Sukai Huang},
 \textbf{Chenyuan Zhang},
 \textbf{Fucai Ke},
 \textbf{Zhixi Cai},
\\
 \textbf{Gholamreza Haffari},
 \textbf{Lizhen Qu},
 \textbf{Hamid Rezatofighi},
\\
\\
 Faculty of Information Technology, Monash University
 \\
 \small{
   \textbf{Correspondence:} \href{mailto:sukai.huang@monash.edu}{sukai.huang@monash.edu}
 }
}

\begin{document}
\maketitle
\begin{abstract}
Instruction granularity is an important yet poorly controlled variable in language-guided embodied AI. Existing benchmarks typically pair each task with a single static instruction, making it difficult to study how agent behavior changes when the same task is described at different levels of detail. We introduce \textsc{Mini-BEHAVIOR-Gran}, a new benchmark for controlled studies of instruction granularity that extends \textsc{Mini-BEHAVIOR} with multiple instruction variants per task, ranging from high-level goal descriptions to step-by-step guidance. Using this benchmark, we compare four candidate metrics for cross-task granularity quantification: token count, entity count, action-verb count, and \emph{planning-width}, and find that width correlates most consistently with agent performance. Using width to organize training and evaluation further reveals a non-monotonic U-shaped relationship between instruction granularity and performance, with peaks at both fine and coarse extremes. Further analysis suggests that the coarse-granularity performance rebound is associated with shallow grounding, where agents learn vision-dominant policies\footnote{Code and data will be released upon publication.}.
\end{abstract}

\section{Introduction}
\label{sec:introduction}

Language-guided embodied agents learn a policy $\pi(a|v,l)$ that maps visual observations $v$ and language instructions $l$ to executable actions $a$. Recent progress has largely focused on model architectures, most notably the Vision-Language-Action (VLA) paradigm \citep{DBLP:conf/corl/KimPKXB0RFSVKBT24,pmlr-v305-black25a,shukor2025smolvla} and its advanced variants that adapt \emph{world models} for future frame prediction ($v_t \mapsto v_{t+1}$) \citep{wu2026pragmaticvlafoundationmodel,DBLP:conf/cvpr/ZhaoLKFZWLMHFHL25,bi2025motusunifiedlatentaction} or \emph{Chain-of-Thought} (CoT) reasoning for intermediate planning \citep{ye2025vlar1enhancingreasoningvisionlanguageaction,lu2025thinkbot,yin2025deepthinkvla}. Yet the impact of \emph{instruction granularity} remains underexplored. Instruction granularity refers to the level of detail provided in $l$ to guide agent decision-making \citep{arumugam2017accurately}, and it is typically an uncontrolled variable in current datasets and evaluations. It matters because varying granularity fundamentally alters the cross-modal alignment difficulty and the latent planning burden imposed on the agent. Understanding this impact is crucial not only for optimizing instruction design during training, but also for ensuring model robustness across diverse human instruction styles encountered at test time.

\begin{figure}
    \centering
    \includegraphics[width=\linewidth]{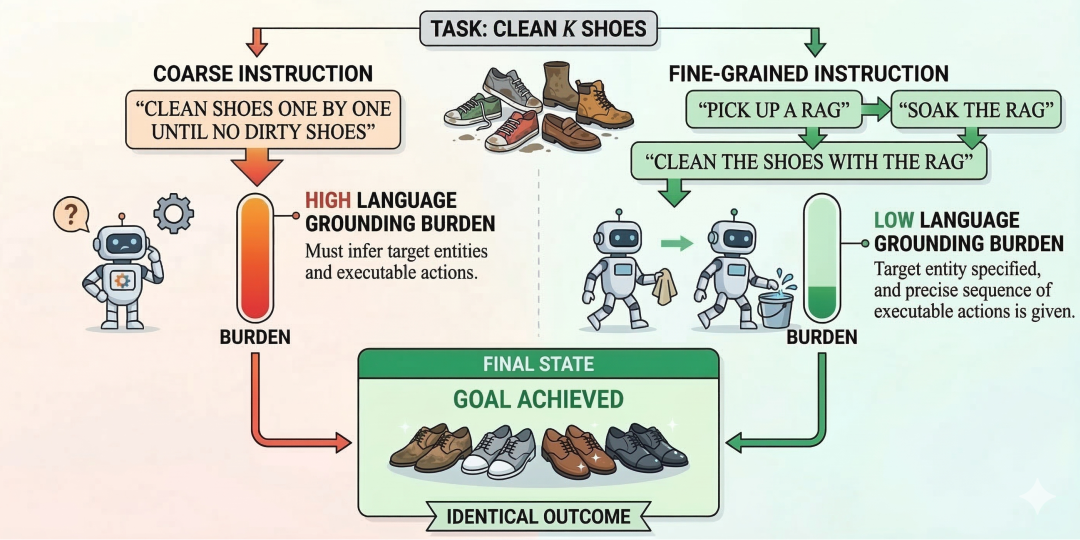}
    \caption{For the same task (cleaning $k$ shoes), a coarse instruction requires the agent to infer target entities and executable actions. In contrast, a fine-grained instruction provides more detailed steps, significantly reducing the language grounding effort to reach the goal.}
    \label{fig:running_example_init}
\end{figure}

A systematic investigation of instruction granularity has been hindered by two key bottlenecks. First, existing embodied AI benchmarks lack controlled variation in instruction granularity. Typically, each task instance is paired with a single, static instruction, making it difficult to study how an agent's performance scales as instructions shift from fine-grained (e.g., detailed step-by-step) to coarse (e.g., high-level goal descriptions). For example, popular benchmarks such as \textsc{LIBERO} and \textsc{BEHAVIOR-1K} provide only fixed or high-level instructions. Even recent suites like \textsc{LIBERO-Plus} \citep{fei2025liberoplus}, which aim to increase diversity, primarily vary visual conditions (e.g., lighting, sensor noise) while keeping the language input fixed.

Second, and perhaps more fundamentally, the field lacks a well-accepted and formalized metric for quantifying \emph{instruction granularity}. Since granularity represents the density of guidance provided for decision-making, an ideal metric should directly reflect the complexity of grounding language into executable actions. With the advent of Large Language Models (LLMs), a seemingly natural proxy is to use the hierarchical depth of an instruction, i.e., its specific level within a decomposed plan, as a measure of granularity \citep{gao2025vlaos}. However, such a metric fails to support meaningful cross-task comparisons because it conflates instructional detail with the task's inherent complexity. Different tasks exhibit fundamentally different internal structures: some involve interleaved subgoals that resist clean hierarchical decomposition (e.g., the Sussman anomaly; \citealp{sussman1975computer}), whereas others can be arbitrarily elongated into superficial substeps. Consequently, hierarchical levels cannot establish a consistent granularity scale across diverse tasks, thereby undermining their utility for investigating the general impacts of granularity on language grounding.

To enable systematic study of how instruction granularity shapes language-guided embodied agents, we introduce \textsc{Mini-BEHAVIOR-Gran}, an extension of \textsc{Mini-BEHAVIOR} \citep{jin2023minibehavior} that provides multiple instruction variants per task, spanning from high-level goal description to step-by-step instructions. Leveraging Minigrid's engine \citep{chevalier-boisvert2023minigrid}, we construct 803 unique problem instances across 20 domains, yielding 10,253 trajectory episodes, each paired with dynamic instructions that evolve as the agent progresses through the task. 
Within this controlled setting, we compare four metrics for quantifying granularity: token count, entity count, action-verb count, and \emph{width}. Unlike \emph{decomposition depth}, \emph{width} measures the max state features jointly tracked per instruction, thereby enabling cross-task comparability of planning complexity~\citep{DBLP:conf/aaai/BonetFG19,DBLP:journals/jair/DrexlerSG24}.

Our study shows that \emph{width} correlates more consistently with agent performance across task horizons and VLA variants than other metrics, serving as a better quantitative proxy for instruction granularity. Organizing training by width reveals a U-shaped pattern. Probing this effect via instruction predictability ($P(l|v)$) suggests that the rebound at coarse end is associated with shallow grounding, where agents prioritize single-modality cues and shift toward vision-dominant policies.

\paragraph{Contribution.} We make the following contributions: \textbf{(1)} Introduce \textsc{Mini-BEHAVIOR-Gran}, a benchmark for controlled study of instruction granularity. \textbf{(2)} Demonstrate that \emph{width} is the most consistent cross-task proxy among the metrics we evaluate. \textbf{(3)} Unveil a non-monotonic U-shaped relationship between granularity and agent performance; and \textbf{(4)} Provide evidence that the coarse-granularity performance rebound is linked to shallow grounding.

\section{\textsc{Mini-Behavior-Gran}: A Controlled Benchmark for Instruction Granularity}

Following prior work on varying granularities and abstraction in robot instructions \citep{DBLP:journals/arobots/ArumugamKGWRWT19}, we define \emph{instruction granularity as the level of details and abstraction with which a language instruction specifies a task. In embodied settings, instruction granularity determines the latent planning complexity that an agent must resolve to map instructions into executable actions.}

Reiterating the example in \Cref{fig:running_example_init}, we note that a coarse instruction (e.g., ``clean the shoes'') leaves the agent with a large language grounding gap, as it must infer the target entities (e.g., which shoes to clean) and the executable actions (e.g., how to clean). In contrast, a fine-grained instruction provides detailed steps that significantly reduce the language grounding effort to fulfill the same task. In this section, we describe how we construct \textsc{Mini-BEHAVIOR-Gran} to systematically vary instruction granularity for the same task.

\subsection{Shared Feature Space and Symbolic Layer}
Many embodied AI simulators expose symbolic representations alongside raw pixel observations for task specification, progress monitoring, or success evaluation \citep{shridhar2020alfred, puig2020watchandhelp, li2022igibson, DBLP:conf/corl/0002ZWGSMWLLSAH22}. We leverage this symbolic backbone to define a shared feature space $\Phi$ common across all tasks, representing each environment state $s$ as a feature vector $\phi(s) = (\phi_1(s), \ldots, \phi_n(s))$. Features in $\Phi$ include both Boolean variables (e.g., $H_r$: holding the rag; $S_r$: the rag is soaked) and numerical variables (e.g., $p_r$: distance to the rag; $N_f$: number of uncleaned shoes). This shared feature space provides the underlying vocabulary for representing instructions, which in turn allows us to quantify instruction granularity across tasks.

\subsection{Rule-Based Instruction Representation}
Building on the shared feature space $\Phi$, we construct each instruction as a set of \emph{``if-then'' rules} that specify desired state progression. This choice is motivated by the fact that sequential decision-making tasks in automated planning are commonly modeled in terms of state variables and transitions with preconditions and effects, following the STRIPS formalism \citep{haslum2019introduction}. Thus, rule-based instructions explicitly encode the latent planning complexity an instruction imposes, as agents must determine how to transition from states satisfying a rule's precondition to those satisfying its effect. Concretely, an instruction in this study is a set of rules $\mathcal{R} = \{r_1, \ldots, r_m\}$, where each rule $r_i$ is of the form $C_{r_i} \mapsto E_{r_i}$:

\begin{itemize}[leftmargin=*]
    \item $C_{r_i}$ (the \emph{condition}) is a conjunction of feature constraints. For a Boolean feature $p$, a constraint is $p$ or $\lnot p$; for a numerical feature $n$, it is $n = 0$ or $n > 0$.
    \item $E_{r_i}$ (the \emph{effect}) is a conjunction of feature changes. For a Boolean feature $p$, an effect is $p$, $\lnot p$, or $p?$ (allowing any change); for a numerical feature $n$, it is $n\downarrow$ (decrease), $n\uparrow$ (increase), or $n?$ (any change).
\end{itemize}

For each task, we begin by manually constructing a finest-grained rule set $\mathcal{R}^{(0)}$ that decomposes the task into atomic steps, i.e., each rule typically can be fulfilled by a single action. We use the task of \emph{cleaning k shoes} as a running example. \Cref{tab:shoe_features} lists the relevant features from the shared space $\Phi$ used in this task.

$\mathcal{R}^{(0)}$ contains over a dozen atomic rules. For brevity, we show representative rules from each phase (full set in \Cref{app_sec:prompt_examples}).

\begin{quote}
\scriptsize
\textbf{Representative rules from $\mathcal{R}^{(0)}$ (Cleaning Shoes).}
\begin{align*}
\{\neg H_r,\, p_r > 0\} &\mapsto \{p_r \downarrow\} && \text{(approach rag)}\\
\{H_r,\, \neg T_s,\, p_s = 0\} &\mapsto \{T_s\} && \text{(turn on tap)}\\
\{H_r,\, T_s,\, p_s = 0\} &\mapsto \{S_r\} && \text{(soak rag)}\\
\{S_r,\, N_f > 0,\, p_f > 0\} &\mapsto \{p_f \downarrow\} && \text{(approach shoe)}\\
\{S_r,\, N_f > 0,\, p_f = 0\} &\mapsto \{N_f \downarrow,\, \neg H_r\} && \text{(clean and drop)}\\
\{N_f = 0,\, \neg O_r,\, H_r\} &\mapsto \{\neg H_r,\, O_r\} && \text{(put away rag)}
\end{align*}
\end{quote}

\begin{table}[t]
\centering
\scriptsize
\begin{tabular}{ll}
\toprule
\textbf{Feature} & \textbf{Description} \\
\midrule
$H_r$ & holding the rag (Boolean) \\
$p_r$ & distance to the nearest rag (numerical) \\
$p_s$ & distance to the sink (numerical) \\
$p_f$ & distance to the nearest uncleaned shoe (numerical) \\
$N_f$ & number of uncleaned shoes (numerical) \\
$T_s$ & tap (sink) is turned on (Boolean) \\
$S_r$ & rag is soaked (Boolean) \\
$C_i$ & shoe $i$ is cleaned (Boolean) \\
$O_r$ & rag is on the floor (Boolean) \\
\bottomrule
\end{tabular}
\caption{Features used in the Cleaning Shoes task.}
\label{tab:shoe_features}
\end{table}

\begin{figure}[t]
    \centering
    \includegraphics[width=\linewidth]{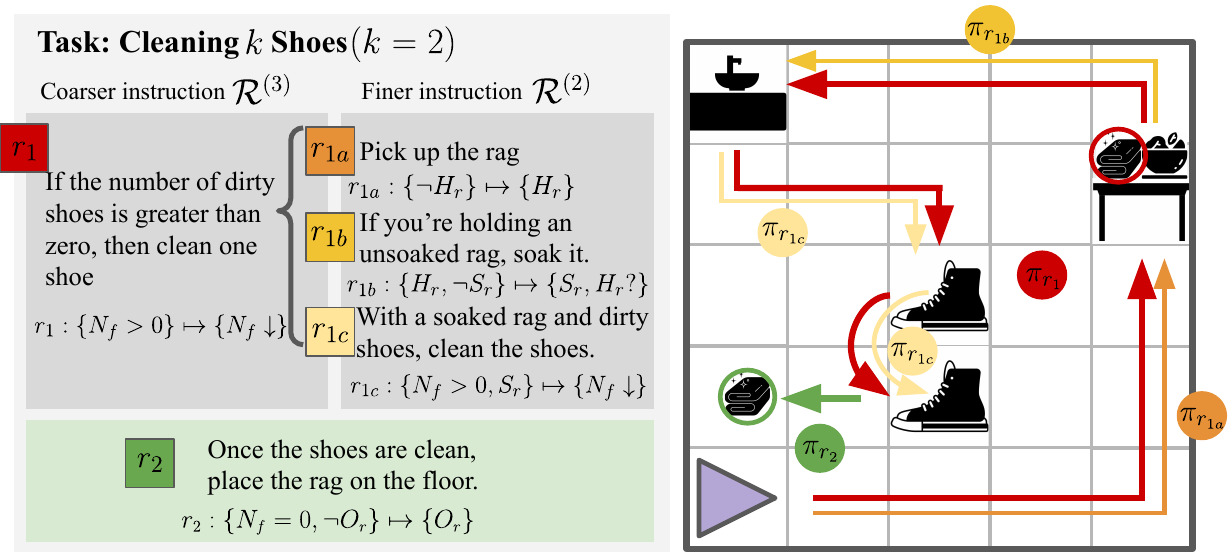}
    \caption{A running example of constructing coarse-grained instructions by merging atomic rules into macro-rules. Features $H_r$ and $S_r$ are omitted as they can be consumed entirely within the subplan chunk.}
    \label{fig:running_example}
\end{figure}

A coarse rule is not created by mechanically concatenating the preconditions and effects of its constituent atomic rules. Instead, we treat each merge as an abstraction of a contiguous subplan segment into a macro-rule. The macro-rule exposes only the \emph{net input/output interface} of that segment: its condition specifies when the segment applies, and its effect captures the overall progress achieved. Features that can be consumed entirely within the segment are thus omitted from the expression. As shown in \Cref{fig:running_example}, the $\mathcal{R}^{(2)}$ rules for picking up the rag ($r_{1a}$), soaking it ($r_{1b}$), and cleaning one shoe ($r_{1c}$) form a coherent subplan $r_1$ that abstracts into a single coarse rule in $\mathcal{R}^{(3)}$, which captures the net progress of cleaning one shoe while hiding the intermediate features $H_r$ and $S_r$. 

After each merge, we verify the semantic validity of the resulting coarse rules through expert review and use a planning solver to ensure every $\mathcal{R}^{(k)}$ is well-formed: sequentially reachable and goal-terminating \citep{DBLP:conf/aaai/BonetG21}.

\begin{figure*}[t]
    \centering
    \includegraphics[width=0.9\linewidth]{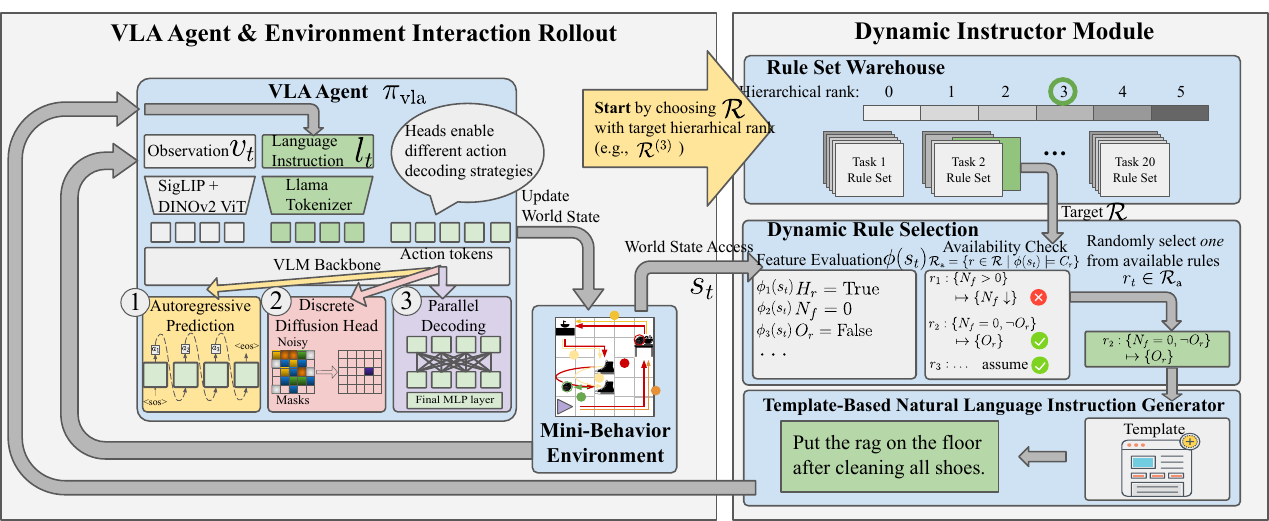}
    \caption{\textsc{Mini-BEHAVIOR-Gran} Experimental Pipeline. We evaluate the impact of instruction granularity across three VLA action-decoding strategies. During inference time, a dynamic instructor evaluates the current world state to decide available rule set $\mathcal R_a$ and active rule $r_t$. A rule-based natural language instruction $l_t$ is generated accordingly. The process iterates until completion or max steps.}
    \label{fig:experiment_illu}
\end{figure*}

\subsection{Dynamic Instruction Generation}

To communicate these rules to the agents, we translate each rule into a natural-language sentence via a template-based generator. The templates verbalize the feature conditions and desired effects in each rule. For example, $\{\hlA{$H_r$}, \hlB{$\neg S_r$}\} \mapsto \{\hlC{$S_r$}\}$ is realized as ``If you are \hlA{holding the rag} but \hlB{the rag is not yet soaked}, \hlC{soak the rag}.''

The benchmark supports \emph{dynamic} instruction generation that updates with task progress. As illustrated in \Cref{fig:experiment_illu}, during both training and evaluation, an instructor module monitors the simulator state at each timestep, computes the set of applicable rules $\mathcal{R}_a$, and determines which rule $r_t$ to activate. This setup ensures that the agent always receives state-relevant guidance, in contrast to existing benchmarks \citep{DBLP:conf/corl/0002ZWGSMWLLSAH22, fei2025liberoplus} that provide only a single static instruction. 

\begin{table}[t]
\centering
\scriptsize
\begin{tabular}{lrrrr}
\toprule
\textbf{Rank} & \textbf{N} & $\overline{\text{Len}}$ & $\textbf{U}_{\text{tok}}$ & $\overline{\text{Plan}}$  \\
\midrule
0 & 51,474 & $40.58 \pm 10.59$ & 145 & $83.43 \pm 53.66$  \\
1 & 27,837 & $38.19 \pm 11.53$ & 139 & $85.81 \pm 56.91$  \\
2 & 14,997 & $33.51 \pm 11.17$ & 253 & $93.68 \pm 57.24$  \\
3 & 6,816  & $34.09 \pm 13.99$ & 171 & $120.26 \pm 51.39$  \\
4 & 1,691  & $41.78 \pm 22.67$ & 112 & $169.32 \pm 37.71$  \\
5 & 923    & $33.31 \pm  7.68$ & 85  & $146.70 \pm 23.56$  \\
\bottomrule
\end{tabular}
\caption{Distribution of instruction variants across granularity ranks. \textbf{N}: number of instruction instances; $\overline{\text{Len}}$: mean instruction length in words; $\textbf{U}_{\text{tok}}$: number of unique tokens; $\overline{\text{Plan}}$: mean plan length in steps.}
\label{tab:instruction_stats_main}
\end{table}

\subsection{Benchmark Analysis}
Table~\ref{tab:instruction_stats_main} summarizes the scale and coverage of \textsc{Mini-BEHAVIOR-Gran}. 
Across 20 domains, we define a shared feature space $\Phi$ with 166 grounded features. 
Task-specific feature usage ranges from only 2 features in simple tasks such as \emph{Opening Packages} to 29 in long-horizon tasks such as \emph{Cleaning Up the Kitchen}. 
Despite their structured origin, the resulting instructions are of moderate linguistic complexity, averaging 40.2 words in length (see \Cref{app_sec:ref_plan_gen} for details).

\section{Quantifying Instruction Granularity}
As highlighted in \Cref{sec:introduction}, a second key bottleneck for 
systematic study of instruction granularity is the absence of a 
well-accepted metric that quantifies it in a way that reflects 
language grounding difficulty and remains comparable across 
diverse tasks. A seemingly natural candidate is 
\emph{decomposition depth}: the level at which an instruction 
sits within a plan hierarchy. However, decomposition depth 
cannot serve as a consistent cross-task metric, because the 
number of attainable levels is determined by each task's 
internal structure. 

\subsection{Candidate Metrics}

To this end, we evaluate four candidate metrics, each corresponding to a different hypothesis about which properties of an instruction plausibly reflect its granularity and associated grounding burden. 
\emph{Token Count} measures \emph{verbal elaboration}: finer-grained instructions often require more words to spell out procedural details.
\emph{Entity Count} measures \emph{referential grounding load}, since mentioning more objects or locations requires the agent to ground a larger set of task-relevant referents. 
\emph{Action-Verb Count} measures \emph{procedural complexity}, as finer-grained instructions tend to specify more individual actions. 
While these three metrics capture surface-level properties, our fourth metric, \emph{Width}, captures a more structural notion of planning burden.

\subsection{Instantiating Width as a Metric}

Among the candidate metrics we consider, \emph{width} is the least surface-oriented. Originating from width-based planning \citep{DBLP:conf/ecai/LipovetzkyG12,DBLP:conf/aaai/LipovetzkyG17}, for a rule $r_i$, its width $w(r_i)$ is the maximum number of features that must be simultaneously tracked along an optimal transition from a state satisfying $C_{r_i}$ to one satisfying $E_{r_i}$. This differs from simply counting the number of features mentioned in the rule's condition or effect, since these features need not all be coordinated concurrently. We instantiate instruction-level width by aggregating over its constituent rules $w(\mathcal{R}) = \max_{r_i \in \mathcal{R}} w(r_i)$. A special case of $w=0$ occurs for situations where the desired state progression can be achieved in zero or one step \citep{bonet2024general}; this will appear in our experimental analysis. 

\begin{table}[t]
\centering
\tiny
\begin{tabularx}{\columnwidth}{l X c}
\toprule
\textbf{State Progression} & \textbf{Concurrent Features Tracked} & \textbf{Width} \\ \midrule
Initial State & $[\langle \rangle]$ & 0 \\
Navigate to rag & $[\langle p_r = X \rangle, \ldots, \langle p_r = 0 \rangle]$ & 1 \\
Pick up rag & $[\langle p_r = 0 \rangle, \langle H_r \rangle]$ & 1 \\
Navigate to sink & $[\langle H_r, p_s = X \rangle, \ldots, \langle H_r, p_s = 0 \rangle]$ & 2 \\
Turn on tap (sink) & $[\langle H_r, p_s = 0  \rangle,  \langle T_s \rangle]$ & 2 \\
Soak rag & $[\langle H_r, T_s, p_s = 0  \rangle, \langle S_r \rangle]$ & \textbf{3} \\
Navigate to Shoe 1 & $[\langle H_r, S_r, p_{f1} \downarrow \rangle, \ldots,] $ & \textbf{3} \\
Clean Shoe 1 & $[\langle H_r, S_r, p_{f1} = 0 \rangle, \langle C_1 \rangle]$ & 3 \\
Navigate to Shoe 2 & $[\langle H_r, S_r, C_1, p_{f2} \downarrow \rangle, \ldots ]$ & \textbf{4} \\
Clean Shoe 2 & $[\langle H_r, S_r, C_1  , p_{f2} = 0 \rangle, \langle C_1, C_2   \rangle]$ & 4 \\
Put rag on floor & $[\langle C_1, C_2, H_r \rangle, \ldots, \langle C_1, C_2, O_r \rangle]$ & 3 \\ \bottomrule
\end{tabularx}
\caption{State progression and width for the ``Cleaning $k$ Shoes'' ($k=2$). Width scales as $w = k+2$.}
\label{tab:task_width_scaling}
\end{table}

\Cref{tab:task_width_scaling} illustrates width computation for a two-shoe cleaning task.  Approaching the rag ($p_r \downarrow \to p_r=0$) requires tracking only $p_r$ ($w=1$). Obtaining the rag at $p_r=0$ ($\to H_r$) also needs only $p_r$ ($w=1$). Then, maintaining $H_r$ while progressing raises width to $2$. Soaking the rag requires simultaneously maintaining three features: at sink ($p_s=0$), holding rag ($H_r$), and tap on ($T_s$), yielding $w=3$. Cleaning the first shoe ($C_1$) requires tracking $H_r$, $S_r$, and $p_f$ ($w=3$). Cleaning the second shoe adds maintaining $C_1$, increasing width to $4$. We acknowledge that $w$ is not a direct measure of internal grounding difficulty, since learning-based embodied agents operate as black-box decision-makers rather than explicit width-based search. However, $w$ provides a cross-task proxy for the latent planning demands. We thus view it as a planning-inspired measure of granularity.

\section{Experiments}

\subsection{Experimental Setup}

\paragraph{Data.}
We use \textsc{Mini-BEHAVIOR-Gran}, which contains 20 long-horizon household tasks defined over a shared feature space $\Phi$. 
For each task, the benchmark provides 50 training and 10 evaluation instances, each paired with reference trajectories generated by the \textsc{BFWS} planning solver \citep{DBLP:conf/aaai/LipovetzkyG17}.

\paragraph{Action-Decoding Archetypes.}
The primary architectural difference among recent VLA models lies in their action-decoding strategy \citep{zhong2025survey}. 
We therefore adopt a unified VLM backbone and instantiate three representative action-decoding paradigms (see the left panel of \Cref{fig:experiment_illu}): 
(1) \textbf{Autoregressive} (AR), exemplified by RT-2 \citep{DBLP:conf/corl/ZitkovichYXXXXW23}, OpenVLA \citep{DBLP:conf/corl/KimPKXB0RFSVKBT24}, and related models \citep{DBLP:journals/tmlr/ReedZPCNBGSKSEBREHCHVBF22,DBLP:conf/icml/HuangYMLLW0ZJ024,DBLP:conf/icra/ONeillRMGPLPGMJ24}, which predict actions sequentially via causal attention; 
(2) \textbf{Discrete Diffusion}, which predicts action chunks through a diffusion process \citep{pmlr-v305-black25a,bjorck2025gr00t,shukor2025smolvla,liang2025discrete}; and 
(3) \textbf{Parallel Decoding}, which uses bidirectional attention for single-pass action prediction \citep{kim2025fine,DBLP:conf/iros/SongCDZZZGLWWML25,yin2025deepthinkvla,wang2025bitvla,xiao2025ava}. 
Implementation details are given in \Cref{app_sec:implementation_details}.

\paragraph{Two-Stage Experimental Protocol.}
\label{sec:experimental_protocol}
We conduct experiments in two stages. 
In \textbf{Stage I} (\Cref{sec:rq1_metric}), we keep the training distribution fixed and train all models under a mixed-granularity regime that samples from all available instruction variants. 
This stage is used only to compare candidate granularity metrics by how consistently they correlates with agent Success Rate (SR) across tasks, task horizons, and action-decoding architectures. 
In \textbf{Stage II} (\Cref{sec:rq2_u_shape}--\Cref{sec:rq3_u_shape}), after identifying which is the most consistent metric, we use it to partition instruction variants into \emph{Fine}, \emph{Medium}, and \emph{Coarse} groups and study how different granularity mixtures affect learning and generalization. 
Specifically, we consider seven training settings: \textbf{Pure F/M/C} ($1\!:\!0\!:\!0$), (2) \textbf{Centroid} ($1\!:\!1\!:\!1$): uniformly sample from all three classes; (3) \textbf{Axial F/M/C} ($3\!:\!1\!:\!1$): emphasizes one class with some exposure to the others.

\subsection{RQ1: Which metric best captures instruction granularity?}
\label{sec:rq1_metric}
We first ask which candidate metric most consistently captures instruction granularity in a way that it reflects embodied language grounding difficulty across tasks. We train all models under a balanced granularity regime and evaluate the Pearson correlation between each candidate metric and agent SR.
A useful cross-task metric should exhibit a stable relationship with performance across tasks of different sequential burdens.

\begin{figure}[t]
    \centering
    \includegraphics[width=\linewidth]{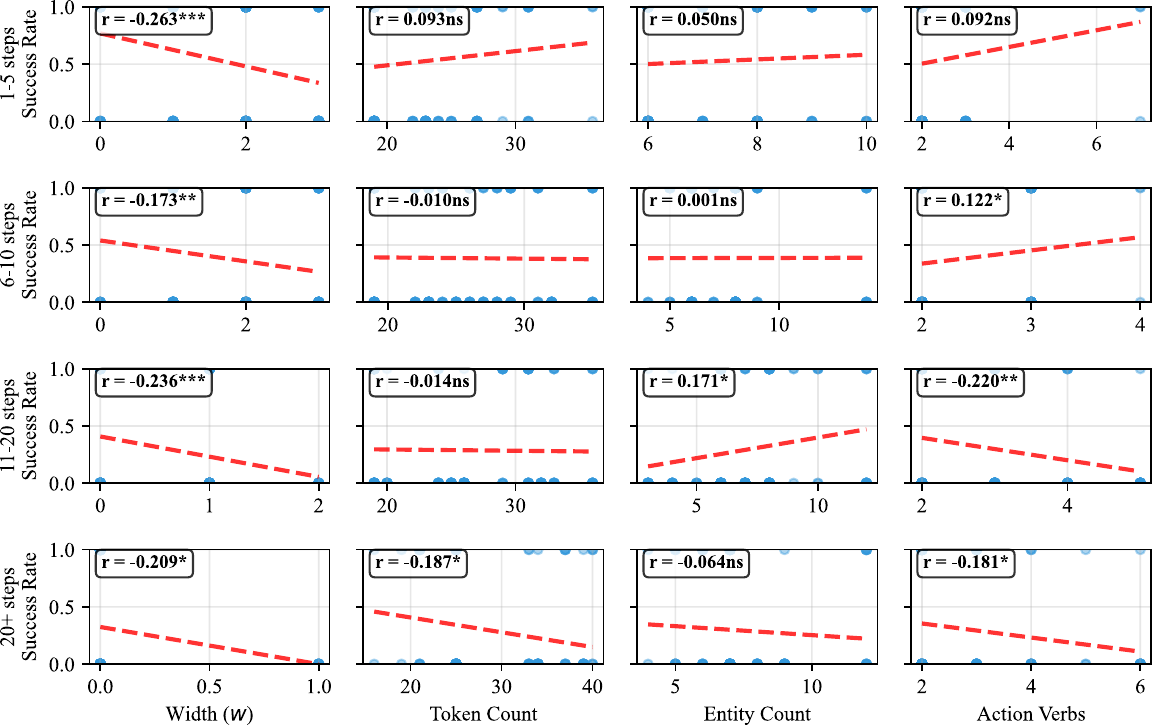}
    \caption{Metric-success correlation by task-horizon, with \emph{width} showing consistent negative correlations.}
    \label{fig:q8_scatter_by_horizon}
\end{figure}

\paragraph{Why partition by instruction horizon?}
A critical confound in evaluating instruction granularity is sequential burden. 
A natural choice would be to stratify examples by the total number of primitive actions in the expert trajectory. 
However, this is not the most relevant notion of difficulty in our setting. 
In \textsc{Mini-BEHAVIOR-Gran}, agents do not ground a single static instruction over the entire trajectory; instead, the instructor issues a sequence of rule-conditioned instruction updates as execution unfolds. 
The more relevant notion of sequential burden is therefore the \emph{instruction horizon}, defined as the number of rule calls issued during execution. 
We thus partition metric--SR associations evaluation instances by instruction horizon.

\paragraph{Width vs.\ surface-level metrics.}
We compare between token count, entity count, action-verb count, and width. 
\Cref{fig:q8_scatter_by_horizon} show that width is the only metric that exhibits a statistically significant and consistently negative correlation with SR across all instruction-horizon bins (e.g, $-0.263$ in the 1-5 steps bin).
In other words, as width increases, agent success decreases in a stable manner regardless of sequential burden. 
By contrast, the other three metrics are notably less consistent. 
Token count is insignificant in the first three bins and becomes significantly negative only in the longest-horizon bin. 
Entity count is mostly insignificant and even becomes weakly positive in the 11--20 bin. 
Action-verb count reaches significance in several bins, but its direction flips across horizons, indicating that it does not provide a stable notion of granularity across tasks. We thus use width to organize training in the remainder of the paper.

\begin{figure}[t]
    \centering
    \includegraphics[width=0.75\linewidth]{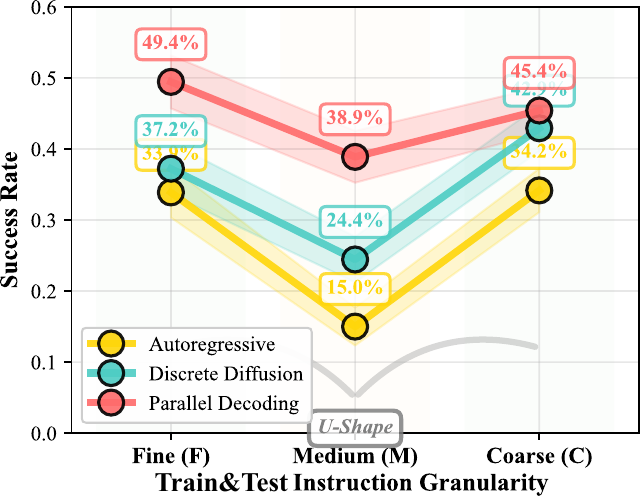}
    \captionof{figure}{In-Distribution (matched train/test) SR across granularities (i.e., train Fine $\rightarrow$ test Fine, etc). A non-monotonic U-shaped trend appears.}
    \label{fig:q1_learning_curves}
\end{figure}

\begin{figure*}[t]
    \centering
    \begin{minipage}[t]{0.40\textwidth}
      \centering
    \includegraphics[width=\linewidth]{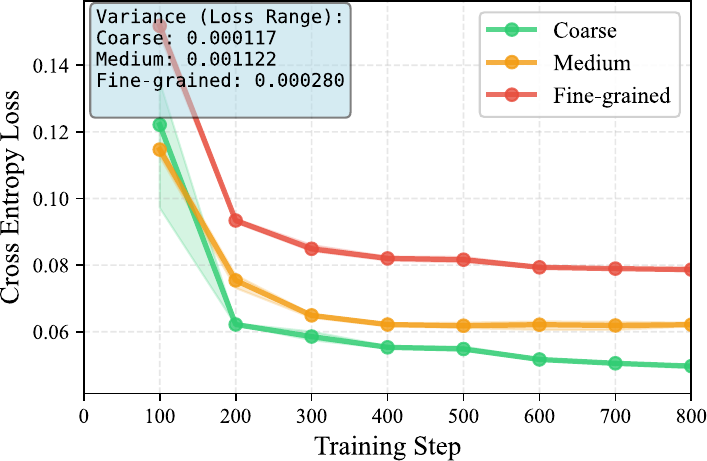}
    \captionof{figure}{VLM instruction-prediction Cross Entropy (CE) loss across granularities and 3 seeds. Larger CE loss indicates lower $P(l \mid v)$.}
    \label{fig:cross_entropy_loss_comparison}
    \end{minipage}\hfill
    \begin{minipage}[t]{0.56\textwidth}
        \centering
        \includegraphics[width=\linewidth]{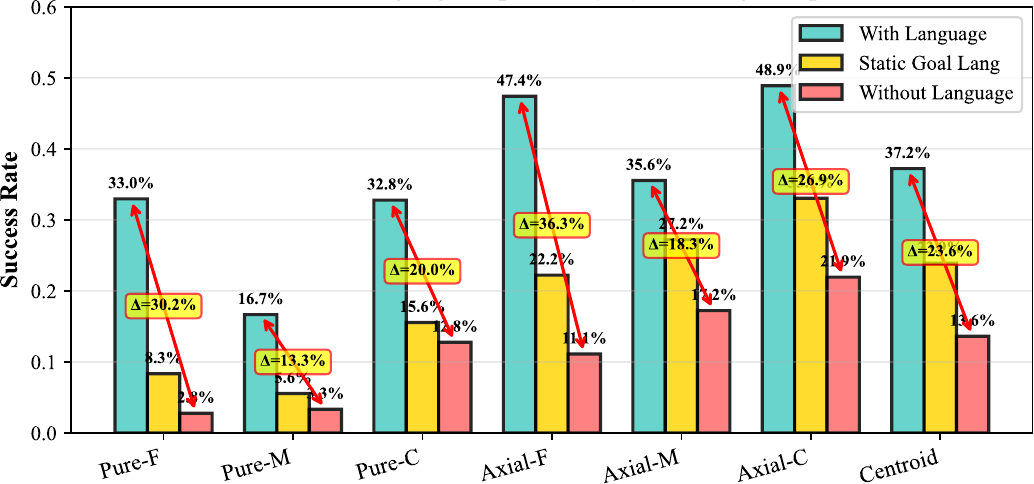}
        \captionof{figure}{Smaller SR gaps indicate a shift towards shallow grounding. Under semantic perturbation, the Pure-C setting exhibits tiny variance ($2.8\%$) between \textit{w/ goal lang} and \textit{w/o lang}.}
        \label{fig:sr_diff_by_training_setup}
    \end{minipage}
\end{figure*}

\begin{figure*}[t]
    \centering
    \noindent
    \begin{minipage}[b]{0.38\textwidth}
    \centering
    \small
    \resizebox{\textwidth}{!}{
    \begin{tabular}{lccc}
    \toprule
    \textbf{Model} & \textbf{T1} & \textbf{T2} & \textbf{T3} \\
    \midrule
    AR        & 78.1\% (25/32)$^{**}$ & 12.5\% (4/32)        & 65.6\% (21/32)       \\
    Parallel  & 15.6\% (5/32)         & 28.1\% (9/32)        & 15.6\% (5/32)        \\
    Diffusion & 53.1\% (17/32)        & 71.9\% (23/32)$^{*}$ & 75.0\% (24/32)$^{**}$ \\
    \bottomrule
    \end{tabular}
    }
    \captionof{table}{Layer-wise ($N=32$) attention trends from action tokens. \textbf{T1}: lang $\uparrow$ ($F \to M$); \textbf{T2}: lang $\downarrow$ ($M \to C$); \textbf{T3}: vis $\uparrow$ ($M \to C$). $^{*}p < 0.05$, $^{**}p < 0.01$ (Binomial test). Parallel models lack consistent trends, likely due to modality entanglement inherent in bidirectional attention \protect\citep{DBLP:conf/icml/KimSK21}. T2/T3 are not zero-sum as the it includes self-attention. Full table and stats in \Cref{app_sec:full_attention_analysis_table}.}
    \label{tab:attention_trends_main}
\end{minipage}\hfill
    \begin{minipage}[b]{0.60\textwidth}
        \centering
        \small
        \resizebox{\textwidth}{!}{
        \begin{tabular}{lccccc}
        \toprule
        \textbf{Training Setup} & \textbf{Width 0} & \textbf{Width 1} & \textbf{Width 2} & \textbf{Width 3} & \textbf{Width $\ge 4$} \\
        & \textbf{(20 tasks)} & \textbf{(20 tasks)} & \textbf{(19 tasks)} & \textbf{(10 tasks)} & \textbf{(3 tasks)} \\
        \midrule
        Pure-F          & 34.9\% & 16.9\% & 10.8\% & 1.0\%  & 0.0\% \\
        Pure-M          & 12.2\% & 22.0\% & 11.7\% & 0.0\%  & 0.0\% \\
        Pure-C          & 26.3\% & 33.3\% & 32.9\% & 22.9\% & 0.0\% \\ 
        \midrule
        Axial-F         & 50.2\% & 53.3\% & 48.8\% & 34.3\% & 0.0\% \\
        Axial-M         & 44.7\% & 43.5\% & 45.8\% & 29.5\% & 0.0\% \\
        Axial-C         & 55.7\% & 52.2\% & 50.8\% & 37.1\% & 0.0\% \\ 
        \midrule
        Centroid (Mix)  & 42.7\% & 38.4\% & 40.8\% & 28.6\% & 0.0\% \\
        \bottomrule
        \end{tabular}
        }
        \captionof{table}{Success rate (\%) across training setups and test granularities. Numbers in parentheses indicate task coverage at each width. For Pure models, cells where test width was unseen during training measure zero-shot transfer. Transfer is asymmetric: coarse $\rightarrow$ fine succeeds.}
        \label{tab:q3_cross_granularity}
    \end{minipage}
    
\end{figure*}

\subsection{RQ2: How does agent performance vary across instruction granularity?}
\label{sec:rq2_u_shape}

To enable a fair cross-task comparison of instruction granularity effects, we construct relative granularity classes via per-task quantile binning. This approach addresses two key limitations of fixed-width grouping: (i) the inherent variation in task complexity results in varying widths (e.g., \emph{opening\_packages} maxes at $w=1$, while \emph{cleaning\_the\_kitchen} reaches $w=5$). Therefore, for each task, we collect all available instruction widths, sort them, and partition them into F/M/C classes using terciles (or an equivalent quantile split that balances the number of instances per level). When the number of distinct widths is insufficient for a clean three-way split (e.g., only two widths exist), we apply a priority-fill heuristic. This normalization ensures that the F/M/C classes represent relative granularity levels within each task, enabling meaningful cross-task comparisons of how performance varies with instruction granularity.

\Cref{fig:q1_learning_curves} reveals a consistent non-monotonic U-shaped relationship between instruction granularity and agent performance across all three action-decoding archetypes. At the fine extreme, where instructions require minimal latent reasoning, models achieve high success rates. Performance then drops at medium granularity, before rebounding under coarse instructions. This is notable because width tracks latent coordination burden: if granularity affected performance only through increasing planning difficulty, one would expect a monotonic decline rather than a recovery at the coarse end.

\subsection{RQ3: What drives the U-shaped effect?}
\label{sec:rq3_u_shape}

The coarse-granularity rebound is initially counterintuitive: if width reflects latent coordination burden, why does performance improve again when instructions become coarser? We hypothesize that this rebound stems from increasing instruction predictability $P(l|v)$, which makes coarse instructions easier to infer from visual context alone. This high predictability incentivizes the policy to exploit single-modality cues, leading to a dominant reliance on visual features rather than complex cross-modal coordination.

This increased predictability has a formal basis: coarse instructions subsume multiple finer-grained variants. Formally, consider a surjective mapping $f$ from fine-grained labels $l_i$ to coarser ones $c_j$. The conditional probability of a coarse label aggregates its fine-grained constituents: $P(c_j | v) = \sum_{l_i \in f^{-1}(c_j)} P(l_i | v)$. This aggregation makes coarse instructions statistically easier to predict from vision alone.

We confirm this empirically by training a VLM-based instructor (Qwen3-VL-4B with LoRA) to predict instructions from visual observations. As \Cref{fig:cross_entropy_loss_comparison} shows, cross-entropy loss decreases monotonically with coarseness (Fine: 0.078, Medium: 0.063, Coarse: 0.047), a relative drop of $40\%$ from fine to coarse, confirming that $P(l|v)$ indeed increases with coarseness.

We next ask whether this increased predictability changes how much models actually rely on language. Following prior work \citep{fei2025liberoplus}, we use $\Delta \mathrm{SR}$ when language is weakened or removed as a probe of grounding reliance. When full instructions are removed entirely, fine-trained policies suffer a severe $36.3\%$ drop, whereas coarse-trained policies show a much smaller $20.0\%$ gap (\Cref{fig:sr_diff_by_training_setup}). Notably, under reduced informativeness (\textit{w/ goal lang}), the Pure-C policy exhibits only a negligible $2.8\%$ advantage over having no language at all ($15.6\%$ vs. $12.8\%$). This indicates that policies trained on coarse instructions develop shallow grounding: language is not deeply integrated into decision-making.

Table \ref{tab:attention_trends_main} provides a \emph{supplementary diagnostic}, for the shallow-grounding account. Using a layer-wise Binomial test ($H_0=0.5$), we find that Discrete Diffusion models exhibit a shift consistent with greater visual reliance at the coarse end (T2, $p=0.01$; T3, $p<0.01$). AR models show a similar but weaker pattern: a stronger language grounding phase from \emph{F} to \emph{M} granularity (T1: $25/32$, $p \approx 0.001$), followed by a directional shift toward visual reliance at the coarse end (T3: $21/32$). We therefore interpret these results as converging evidence rather than definitive mechanistic explanation. Note that T2 and T3 are not redundant: action tokens can also attend to previous action tokens, so a decrease in language attention does not imply a one-to-one increase in visual attention. 
 
Together, these results reveal a three-regime account of the U-shaped performance pattern. At \textbf{fine granularity}, Low $P(l|v)$ (high CE loss) means language provides unique information. Combined with low-width grounding burden, this drives genuine multimodal alignment for generating actions, yielding high SR. \textbf{Medium granularity} introduces a moderate increase in grounding burden. The model attempts to ground language but struggles with the increased complexity, leading to the lowest performance. At \textbf{coarse granularity}, instructions are too complex to ground effectively, and high $P(l|v)$ incentivizes agents to exploit visual cues. This visual-dominant strategy recovers SR but sacrifices language grounding.

\paragraph{Grounding via Mixed-Granularity Training.}
We hypothesize that increasing the diversity of the training language space should artificially lower average $P(l|v)$, forcing agents to re-engage with text. Our mixed-granularity training confirms this: all mixed-granularity models exhibit substantially larger $\Delta\mathrm{SR}$ gaps than their Pure-X counterparts. Among these, the fine-weighted mixture \emph{Axial-F} emerges as the optimal compromise, achieving strong overall performance while maintaining larger $\Delta\mathrm{SR}$ gaps indicative of language grounding.

\paragraph{Asymmetric Generalization.} We observe an asymmetric transfer pattern (\Cref{tab:q3_cross_granularity}): coarse-trained models successfully zero-shot transfer to fine-grained instructions, but fine-trained models fail on coarse-grained ones. This asymmetry aligns with our $\Delta\mathrm{SR}$ analysis: coarse training induces a visual-dominant policy, hence refining the instruction yields minimal behavioral change. Fine training induces tight coupling with language, making the policy brittle when detailed guidance drops. Also see \Cref{app_sec:additional_experimental_results} for failure mode analysis, showing low width encounters regression failures more frequently, while high width leads to stagnation.

\section{Related Work}
\label{sec:related_work}

\textbf{Instruction Abstraction and Granularity.}
Here, \emph{instruction granularity} refers to the level of detail that affects the latent planning burden of an instruction. This differs from \emph{linguistic abstraction}, defined as representational compression~\citep{Wing2010ComputationalTW,DBLP:journals/tacl/LachmyPMT22,DBLP:conf/iclr/ZhengMCCCLZ24}. Granularity varies independently: ``tidy the room'' is coarse yet non-abstract (underspecified actions, no compression); ``repeat wiping until clean'' is specific yet abstract (prescribes action via looped condition).

While not framed as the term ``granularity'', we acknowledge that a rich body of work has focused on augmenting language specificity and its impact, for example through coarse-to-fine semantic parsing \citep{dong-lapata-2018-coarse,DBLP:conf/ijcai/LiLF020}.
Other work generates compositional instructions via constraint dropout and sub-task recombination \citep{huang-etal-2025-musc}, or construct counterfactual instructions to augment diversity \citep{glossop2025castcounterfactuallabelsimprove}.

Varying instruction detail is often handled via a simple two-part decomposition: high-level plans are converted into sequences of low-level steps before being grounded into actions \citep{arumugam2017accurately, DBLP:conf/wuwnet/Yang0H24,DBLP:conf/icml/ShiIEKPVTWWFLDG25}. Most recent Vision Language Action (VLA) works, however, operate with static language instructions \citep{ma2025surveyvisionlanguageactionmodelsembodied,li2025cogvla}. Some models like InstructVLA and ThinkAct explore having chain-of-thought reasoning or planning steps before action decoding \citep{DBLP:conf/corl/HuangXXCLFZTMCS22, yang2025instructvlavisionlanguageactioninstructiontuning,huang2025thinkactvisionlanguageactionreasoningreinforced}. These hierarchical definitions remain task-specific and qualitative, lacking a unified, cross-task metric to establish shared granularity levels across tasks. Consequently, quantitatively controlled studies on the formal dynamics of instruction granularity remain scarce.

\noindent\textbf{Language Grounding vs. Shortcuts.} 
Prior work highlights a dichotomy between fragile language sensitivity \citep{akula-etal-2022-alfred} and visuo-motor shortcuts \citep{xing2025shortcut, zhang2026restoringlinguisticgroundingvla}. We reconcile this via width ($w$), providing complementary insights into the trade-offs between language grounding and visual reliance based on varying instruction granularity in training distributions.

\section{Conclusion}
\label{sec:conclusion}

We introduce \textsc{Mini-BEHAVIOR-Gran}, the first testbed for studying the impact of instruction granularity. Comparing four metrics, we find \emph{width} most consistently correlates with grounding difficulty. Using width to organize training reveals a U-shaped granularity-performance pattern. We provide a granularity-centric diagnosis of the underlying mechanisms, showing that coarse instructions are more predictable from vision, leading to visual-dominant policies that exploit visual cues at the expense of language grounding. The results also suggest that training with a fine-weighted mixtures of granularities helps mitigate it.

\section{Limitations}
\label{sec:limitations}

\paragraph{Discrete Action Spaces.} We evaluate our framework in environments with discrete action spaces, a common abstraction in language-conditioned embodied AI (e.g., ALFWorld-based agent \citep{xiong-etal-2025-mpo, kim-etal-2025-reflact}) and LLM-based agent systems \citep{DBLP:conf/iclr/QinLYZYLLCTQZHT24, DBLP:conf/nips/PatilZ0G24}, where high-level reasoning is decoupled from low-level execution via skill primitives or API calls. This choice is intentional: it isolates symbolic planning from continuous control noise, allowing us to cleanly attribute observed effects to instruction granularity. However, extending this granularity-aware framework to hybrid discrete-continuous or continuous action spaces remains essential for real-world robotics applications, where low-level motor commands interact with high-level linguistic abstractions.

\paragraph{Scope of Granularity Definition.} We view subgoal decomposition as the primary axis of instruction granularity, as it directly determines the latent planning burden—the central construct we aim to measure. This focus aligns with the classical planning view that hierarchical task decomposition is the fundamental mechanism for reducing search complexity. Other forms of granularity (e.g., vagueness, abstraction via loops or branching) are important but introduce additional complexities that would confound our initial investigation. By first isolating and understanding the effect of subgoal decomposition in a controlled setting, we establish a foundation for future work to incorporate these other dimensions. Our benchmark and findings provide a baseline against which such extensions can be compared.

\paragraph{Practicality of width computation.}
Computing width currently requires access to symbolic state features and optimal plans, which limits its direct applicability to natural language instructions in unstructured environments. We emphasize, however, that width is designed as an analytic metric for controlled studies, not as a deployable measure for real-world settings. Within our benchmark, width enables precise quantification of latent planning burden, allowing us to isolate the effect of instruction granularity and uncover fundamental phenomena such as the U-shaped performance curve and asymmetric transfer. Extending width estimation to realistic settings remains future work and does not affect the validity of our findings.

\bibliography{custom}

\appendix

\section{Use of LLM Statement} 
An LLM was used only for polishing language; the scientific content is entirely original.

\section{Technical Appendix}

\section*{Appendix Table of Contents}
\begin{enumerate}[leftmargin=*]
    \item \textbf{Anticipatory defense on the choice of test environments and metrics} (\Cref{app_sec:ant_defense_test_env}) \\
    Justification for using \textsc{Mini-BEHAVIOR} over alternatives (e.g., iGibson, CALVIN) based on controllability, symbolic abstraction, and isolation of language-grounding challenges.
    \item \textbf{Symbolic Representation of the Embodied Task} (\Cref{app_sec:classical_prob_def})
    Formal definition of states, actions, and planning problems using classical STRIPS/PDDL semantics.
    \item \textbf{Details of the \textsc{Mini-BEHAVIOR} Environment} (\Cref{app_sec:minibehavior_details})
    Overview of scene layout, object types, action space, and perception model.
    \item \textbf{Details of the \textsc{Mini-BEHAVIOR-Gran} Extension} (\Cref{app_sec:minibehavior_gran_details})
    Procedure for generating multi-granularity instructions, details on the feature set, dynamic instructor, and dataset construction.
    \item \textbf{Rule Set and Natural Language Prompt Examples} (\Cref{app_sec:prompt_examples})
    Details of the rule set bank of 20 tasks and sample generated instructions at different granularities.
    \item \textbf{Implementation Details} (\Cref{app_sec:implementation_details})
    Model architecture (Prismatic-7B backbone, SigLIP+DINOv2 visual encoders), training hyperparameters, optimizer settings. VLM Predictor architecture and training details.
    \item \textbf{Additional Experimental Results} (\Cref{app_sec:additional_experimental_results})
    Extended analysis and ablations.
\end{enumerate}

\section{Anticipatory defense on the choice of test environments and metrics}
\label{app_sec:ant_defense_test_env}

In this work, we selected Mini-BEHAVIOR as our testbed. This choice is motivated by several key considerations aligned with our research objectives. Our research objective is not to solve the full stack of robotic embodiment (which includes perception, continuous control, and physics), but specifically to isolate and analyze the impact of instruction granularity on language grounding and long-horizon planning.

Therefore, high-fidelity simulators (e.g., iGibson \citep{li2022igibson}, BEHAVIOR-1K \citep{DBLP:conf/corl/0002ZWGSMWLLSAH22}) introduce significant noise through complex texture rendering and continuous motor control. While realistic, these factors act as confounders. When an agent fails in such environments, it is often difficult to attribute the failure to a lack of linguistic understanding (granularity issues) or a failure in low-level execution (e.g., grasping physics or occlusion). Mini-BEHAVIOR retains the high-level decision complexity of household tasks—long horizons, multi-object interactions, and state dependencies—while abstracting away low-level sensorimotor noise.

In the perspective of fast prototyping and iterative experimentation, we have to also consider the rendering and physics simulation costs. Mini-BEHAVIOR's grid-based discrete control and simplified visual representation allow for rapid data generation and model training. 

Another consideration is the diversity of the task structures so that we can systematically vary instruction granularity across a wide range of scenarios. As such, Crafter \citep{hafner2021crafter}, which focuses on survival in a single domain, lacks the diverse task structures needed to obtain robust insights about instruction granularity. 

Table \ref{tab:env_comparison} provides a detailed comparison of Mini-BEHAVIOR against other prominent benchmarks, highlighting why they were less suitable for this specific investigation.

\begin{table*}[t]
\small
\centering
\caption{Comparison of Embodied AI Environments regarding their suitability for studying Instruction Granularity.}
\label{tab:env_comparison}
\begin{tabular}{@{}l>{\raggedright\arraybackslash}p{1.4cm}c>{\raggedright\arraybackslash}p{1.5cm}>{\raggedright\arraybackslash}p{1.7cm}>{\raggedright\arraybackslash}p{4.9cm}@{}}
\toprule
\textbf{Environment} & \textbf{Control} & \textbf{Horizon} & \textbf{Visuals} & \textbf{Symbolic Abstraction} & \textbf{Limitation} \\
\midrule
\textbf{Mini-BEHAVIOR}(*) & Discrete & Long & Grid & \textbf{Support} & \textbf{Ideal balance:} Supports symbolic state abstraction; diverse task suite; removes control noise while retaining logic complexity. \\
\textbf{Crafter} & Discrete & Long & Cartoon & Partial & Single task suite (survival); partially observable (memory heavy); few object types. \\
\textbf{Meta-World} & Continuous & Short & 3D & No & Continuous control; no symbolic abstraction; focus on multi-task RL/manipulation, not long-horizon planning. \\
\textbf{LIBERO} & Continuous & Short & 3D & No & Robotic arm focus; continuous control; no symbolic abstraction; VLA success rate already saturated (tasks relatively easy). \\
\textbf{RoboTwin 2.0} & Continuous & Short & 3D & No & Continuous control; no symbolic abstraction; dual-arm manipulation focus; hard to segment expert trajectories and construct granular instructions. \\
\textbf{CALVIN} & Continuous & Medium & 3D & No & Hard to design instruction granularity due to continuous action space (though provides diverse language); confounds planning with control difficulties. \\
\textbf{ALFRED} & Discrete & Long & 3D Photo & Support & 3D partially observable; poor infrastructure support for VLA training; high rendering cost. \\
\textbf{iGibson / BEHAVIOR-1K} & Continuous & Long & 3D Photo & Partial & Good for long-horizon tasks, but 3D realistic vision context becomes a confounder. \\
\bottomrule
\end{tabular}
\end{table*}

\section{Symbolic Representation of the Embodied Task}
\label{app_sec:classical_prob_def}

We formalize the Mini-BEHAVIOR environment as a deterministic grid-based symbolic world. This allows us to represent the challenge as a classical planning problem $\mathcal{P} = \langle \mathcal{F}, \mathcal{A}, s_0, G \rangle$, where:
\begin{itemize}[leftmargin=*]
    \item $\mathcal{F}$ is a finite set of propositional fluents representing the state variables of the environment (e.g., object locations, states like \textit{cooked} or \textit{open}).
    \item $\mathcal{A}$ is a finite set of actions available to the agent.
    \item $s_0 \subseteq \mathcal{F}$ is the initial state configuration.
    \item $G \subseteq \mathcal{F}$ is the goal condition that must be satisfied.
\end{itemize}
A state $s \in \mathcal{S}$ is defined as a truth valuation over $\mathcal{F}$, with the state space $\mathcal{S} \subseteq 2^\mathcal{F}$ encompassing all valid combinations of fluent truth values.

\section{Details of \textsc{Mini-BEHAVIOR} Environment}
\label{app_sec:minibehavior_details}

To operationalize our formalism, we developed a \textbf{Universal PDDL Domain Model} (available at \url{minibehavior_gran_domain_model.pddl} in the codebase) capable of supporting all 20 household tasks without modification. This unified model comprises over 1,000 lines of PDDL 2.1 definitions and is fully compatible with the LAMA planner. We highly recommend checking the codebase for the complete domain model.

\paragraph{Search Space Characteristics}
\begin{table}[H]
    \centering
    \small
    \caption{Planning complexity in typical tasks.}
    \label{tab:planning_complexity}
    \begin{tabular}{@{}lc@{}}
        \toprule
        \textbf{Metric} & \textbf{Typical Range} \\
        \midrule
        Objects & 15--40 \\
        Grid Locations & 20--60 \\
        Ground Actions & $10^3$--$10^5$ \\
        Plan Length & 50--200 primitive actions \\
        Planning Time (LAMA) & 0.5--300+ seconds \\
        \bottomrule
    \end{tabular}
\end{table}

\section{Details of \textsc{Mini-BEHAVIOR-Gran} Extension}
\label{app_sec:minibehavior_gran_details}

\textsc{Mini-BEHAVIOR} provides the foundational task skeletons, which we extend in \textsc{Mini-BEHAVIOR-Gran} to support multi-granularity instruction generation. A core component of this extension is the definition of a symbolic feature set used to generate rule-based policy sketches.

\subsection{Feature Categories}

The environment defines two primary categories of features: Boolean State Features and Quantitative Features.

\subsubsection{Boolean State Features}
These features evaluate whether a specific object instance satisfies a predicate. They require grounding (binding to specific object instances, e.g., \texttt{rag-01}).

\begin{table}[t]
\centering
\caption{Boolean State Features}
\label{tab:bool_features}
\resizebox{\columnwidth}{!}{
\begin{tabular}{l|l|l}
\hline
\textbf{Feature} & \textbf{Applicable Objects} & \textbf{Semantics} \\ \hline
\texttt{Holding} & All Items & Is the agent holding the specified object? \\
\texttt{isOpened} & Cabinets, Drawers, Doors & Is the container/door open? \\
\texttt{isToggled} & Sinks, Stoves, Switches & Is the appliance turned on? \\
\texttt{isSoaked} & Rags, Sponges & Is the cleaning tool wet? \\
\texttt{isCleaned} & Shoes, Dishes, Floor & Is the object free of stains? \\
\texttt{isDustFree} & Furniture, Surfaces & Is the object free of dust? \\ \hline
\end{tabular}
}
\end{table}

\subsubsection{Quantitative and Spatial Features}
These features capture distances, counts, and spatial relationships. Crucially, they support temporal comparison allowing conditions such as ``distance decreased'' or ``count dropped to zero.''

\begin{itemize}[leftmargin=*]
\item \textbf{\texttt{DistanceToNearest(type, condition)}}: Measures the Manhattan distance from the agent to the nearest object of a specific type.
\begin{itemize}[leftmargin=*]
\item \textit{Conditions:} \texttt{=0} (adjacent), \texttt{>0}, \texttt{increase}, \texttt{decrease}, \texttt{change}.
\end{itemize}
\item \textbf{\texttt{Count(type, condition, function)}}: Counts the number of objects of a specific type that satisfy a filter function.
\begin{itemize}
\item \textit{Conditions:} \texttt{=0}, \texttt{>0}, \texttt{increase}, \texttt{decrease}.
\end{itemize}
\end{itemize}

\subsection{Count Functions}
A diverse set of counting functions is implemented to support complex logical conditions (e.g., ``all shoes are clean'' is equivalent to \emph{count\_not\_cleaned == 0}).

\begin{table}[t]
\centering
\caption{Available Count Functions for Rule Definitions}
\label{tab:count_funcs}
\resizebox{\columnwidth}{!}{
\begin{tabular}{l|l}
\hline
\textbf{Category} & \textbf{Function Description} \\ \hline
\textbf{State-Based} & \texttt{count\_not\_cleaned}: Objects that are currently stained. \\
& \texttt{count\_not\_soaked}: Cleaning tools that are dry. \\
& \texttt{count\_not\_toggled}: Appliances that are turned off. \\
& \texttt{count\_closed}: Containers/doors that are closed. \\
& \texttt{count\_not\_sliced}: Food items that require slicing. \\ \hline
\textbf{Spatial} & \texttt{count\_not\_onfloor}: Objects not placed on the floor layer. \\
& \texttt{count\_isolated}: Objects with no neighbors of the same type. \\
& \texttt{count\_not\_inside\_target}: Objects not contained within a specific furniture. \\
& \texttt{count\_not\_ontop}: Objects not placed on top of a specific surface. \\
& \texttt{count\_not\_near\_target}: Objects not within 1 step of a target type. \\ \hline
\textbf{Task-Specific} & \texttt{count\_incomplete\_salad}: Plates lacking necessary ingredients. \\
& \texttt{count\_not\_complete\_tables}: Tables lacking specific setups (e.g., candles). \\ \hline
\end{tabular}
}
\end{table}

\subsection{Reference Plan Generation and Task Complexity Analysis} \label{app_sec:ref_plan_gen}

While the original \textsc{Mini-BEHAVIOR} framework provides the high-level skeletons for household activities, it does not provide reference plans by default. Thus, we utilize the \textsc{BFWS} (Best-First Width Search) classical planner to generate reference plans for all 20 tasks in \textsc{Mini-BEHAVIOR-Gran}. These reference plans serve as the ground truth for our agents and allow us to quantitatively categorize the complexity of each task.

\begin{figure*}[t]
    \centering
    \includegraphics[width=\textwidth]{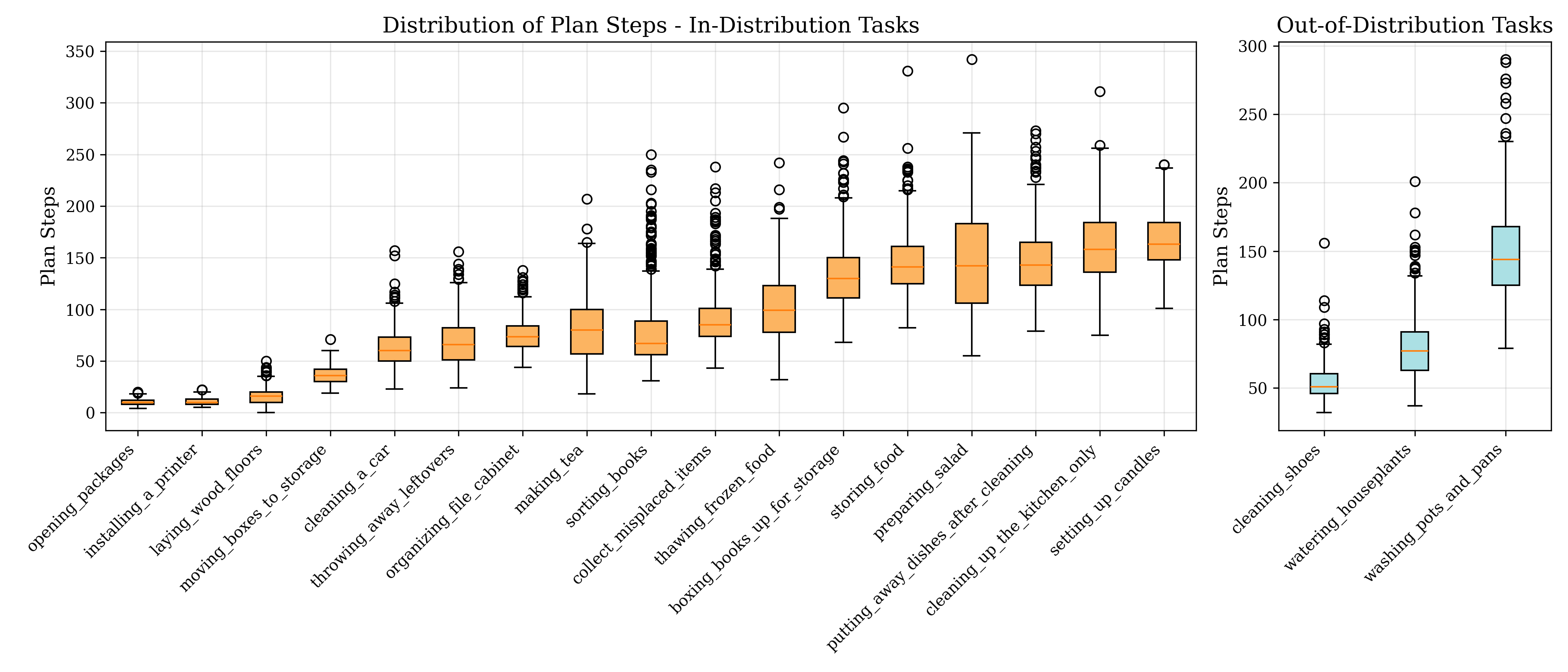}
    \caption{Distribution of reference plan steps required for tasks in \textsc{Mini-BEHAVIOR-Gran} obtained by \textsc{BWFS} classical planner.}
\end{figure*}

\begin{table*}[ht]
\small
\vspace{3mm}
\centering
\label{table:benchmark_task_spec}\scalebox{0.90}{{
\begin{tabular}{>{\centering\arraybackslash}p{4.9cm}||c|c|p{8.4cm}|c}
\toprule[0.4mm]
\cellcolor{mygray}\textbf{Task} & \cellcolor{mygray}\textbf{Split} & \cellcolor{mygray}\textbf{Class} & \cellcolor{mygray}\textbf{Goal Description} & \cellcolor{mygray}\textbf{Mean Steps} \\
\hline \hline
opening\_packages & ID & \cellcolor{green!20}\textbf{short} & Open every package. & 10.2 \\
installing\_a\_printer & ID & \cellcolor{green!20}\textbf{short} & Place a printer on a table and turn it on. & 10.6 \\
laying\_wood\_floors & ID & \cellcolor{green!20}\textbf{short} & Lay every plywood sheet on the floor, each touching at least one other sheet. & 15.9 \\
moving\_boxes\_to\_storage & ID & \cellcolor{green!20}\textbf{short} & Place every carton inside a shelf. & 36.3 \\
cleaning\_a\_car & ID & \cellcolor{green!20}\textbf{short} & Make at least one car dust-free. Place a rag and soap together inside a bucket. & 62.6 \\
throwing\_away\_leftovers & ID & \cellcolor{yellow!25}\textbf{middle} & Throw away every hamburger by placing it in a trash can. & 68.3 \\
organizing\_file\_cabinet & ID & \cellcolor{yellow!25}\textbf{middle} & Put a marker on a table and file every folder and document in a cabinet. & 75.5 \\
making\_tea & ID & \cellcolor{yellow!25}\textbf{middle} & Slice a lemon. Put a teapot on an activated stove. Keep a soaked teabag with the teapot. & 78.7 \\
sorting\_books & ID & \cellcolor{yellow!25}\textbf{middle} & Place every book and every hardback on a shelf. & 78.8 \\
collect\_misplaced\_items & ID & \cellcolor{yellow!25}\textbf{middle} & Place all gym shoes, necklaces, notebooks, and socks on a table. & 90.7 \\
thawing\_frozen\_food & ID & \cellcolor{yellow!25}\textbf{middle} & Place a date label next to a fish. Put every fish next to a sink. Set an olive next to a sink. & 101.7 \\
boxing\_books\_up\_for\_storage & ID & \cellcolor{blue!15}\textbf{long} & Place every book inside a box. & 134.4 \\
storing\_food & ID & \cellcolor{blue!15}\textbf{long} & Store every cookable item inside a cabinet. & 144.6 \\
preparing\_salad & ID & \cellcolor{blue!15}\textbf{long} & Slice all apples and tomatoes; place them on plates. Put every radish and lettuce on plates. Ensure each plate holds at least one cookable item. & 146.7 \\
putting\_away\_dishes\_after\_cleaning & ID & \cellcolor{blue!15}\textbf{long} & Put every plate inside a cabinet. & 147.7 \\
cleaning\_up\_the\_kitchen\_only & ID & \cellcolor{blue!15}\textbf{long} & Put every blender on a countertop. Refrigerate all apples and casseroles. Keep soap next to a sink. Make plates unstained and cabinets dust-free. Place each rag either beside or inside the sink. Store plates and vegetable oil in different cabinets. & 161.0 \\
setting\_up\_candles & ID & \cellcolor{blue!15}\textbf{long} & On every table, place three distinct candles on top. & 165.2 \\
\hline
cleaning\_shoes & OOD & \cellcolor{green!20}\textbf{short} & Leave a towel on the floor and make sure every shoe is unstained and dust-free. & 54.2 \\
watering\_houseplants & OOD & \cellcolor{yellow!25}\textbf{middle} & Water every potted plant until it is soaked. & 79.4 \\
washing\_pots\_and\_pans & OOD & \cellcolor{blue!15}\textbf{long} & Clean every teapot, kettle, and pan so they're unstained, then store each inside a cabinet. & 149.5 \\
\bottomrule[0.4mm]
\end{tabular}}}
\caption{Classification of tasks in \textsc{Mini-BEHAVIOR-Gran} based on the median number of plan steps, as determined by the \textsc{BWFS} planner. Tasks are categorized into short, middle, and long classes to reflect varying planning complexity based on plan length.}
\label{table:benchmark_class_based_on_plan_steps}
\end{table*}

Detailed statistics regarding the specific rule sets used for instruction generation and the resulting natural language statistics are discussed in the next section.

\section{Rule Set and Natural Language Prompt Examples For Different Tasks}
\label{app_sec:prompt_examples}

We present the rank 0 rule set for 10 over 20 household tasks in the Mini-BEHAVIOR-Gran environment. For other levels, please refer to the codebase at \url{mini-behavior-gran/mini_behavior/utils/policy_sketch/}. 

\subsection{Task 1: Laying Wood Floors}

\textbf{Features:}
\begin{itemize}[noitemsep]
    \item $H$: holding an isolated wood (plywood)
    \item $p$: distance to the nearest isolated wood
    \item $t$: distance to an arbitrary (non-held) wood we'll place next to
    \item $n$: number of isolated woods
\end{itemize}

\resizebox{\columnwidth}{!}{$
\begin{aligned}
\{\neg H, p>0, n >0\} &\mapsto \{p \downarrow, t?\} && \text{; move close to isolated wood} \\
\{\neg H, p=0, n >0\} &\mapsto \{H\} && \text{; pick up isolated wood when reachable} \\
\{H, t>0, n >0\} &\mapsto \{t\downarrow\} && \text{; move to some wood location} \\
\{H, n >0, t=0\} &\mapsto \{H?, n \downarrow, p?\} && \text{; drop the isolated wood next to other wood}
\end{aligned}
$}
\subsection{Task 2: Preparing Salad}

\textbf{Features:}
\begin{itemize}[noitemsep]
    \item $H_n$: holding a not-sliceable food
    \item $H_s$: holding a sliceable food
    \item $H_k$: holding a slicer
    \item $U$: number of sliceable food that is not sliced yet
    \item $d_p$: distance to the selected incomplete plate $p$
    \item $d_n, d_s$: distances to nearest not-sliceable / sliceable
    \item $k$: distance to a slicer (knife)
    \item $B$: selected plate $p$ complete bottom salad
    \item $n$: number of incomplete plates
    \item $nn$: number of plates that do not have bottom salad yet
\end{itemize}

\resizebox{\columnwidth}{!}{$
\noindent\textit{Acquire slicer:}
\begin{aligned}
\{U>0 , \neg H_k, k>0\} &\mapsto \{k \downarrow\} && \text{; move to slicer} \\
\{U>0 , \neg H_k, k=0\} &\mapsto \{H_k\} && \text{; hold a slicer} \\
\{U>0 , H_k, d_s > 0\} &\mapsto \{d_s \downarrow\} && \text{; move to sliceable food} \\
\{U>0 , H_k, d_s = 0\} &\mapsto \{U \downarrow, \neg H_k \} && \text{; cut sliceable food} \\
\{U=0 , H_k\} &\mapsto \{\neg H_k\} && \text{; put down slicer}
\end{aligned}
$}
\resizebox{\columnwidth}{!}{$
\noindent\textit{Place bottom salad:}
\begin{aligned}
\{n>0, U=0, nn>0, \neg B, \neg H_n, d_n>0\} &\mapsto \{d_n \downarrow\} \\
\{n>0, U=0,nn>0, \neg B, \neg H_n, d_n=0\} &\mapsto \{H_n\} \\
\{n>0, U=0,nn>0, \neg B, H_n, d_p>0\} &\mapsto \{d_p\downarrow\} \\
\{n>0, U=0, nn>0, \neg B, H_n, d_p=0\} &\mapsto \{B, nn \downarrow, \neg H_n\}
\end{aligned}
$}

\resizebox{\columnwidth}{!}{$
\noindent\textit{Handle sliceable salad:}
\begin{aligned}
\{nn=0, n>0, U=0 , \neg H_s, d_s>0\} &\mapsto \{d_s \downarrow\} && \text{; move to sliceable food} \\
\{nn=0, n>0, U=0 , \neg H_s, d_s=0\} &\mapsto \{H_s\} && \text{; pick up sliceable food} \\
\{nn=0, n>0, U=0 , H_s, d_p>0\} &\mapsto \{d_p \downarrow\} && \text{; move to the plate} \\
\{nn=0, n>0, U=0 , H_s, d_p=0\} &\mapsto \{n \downarrow, \neg H_s\} && \text{; put down sliced food}
\end{aligned}
$}

\subsection{Task 3: Cleaning Up the Kitchen Only}

\textbf{Features:}
\begin{itemize}[noitemsep]
    \item \textit{Holding flags:} $H_b$ (blender), $H_f$ (food), $H_s$ (soap), $H_r$ (rag), $H_p$ (plate), $H_o$ (vegetable oil)
    \item \textit{Distances:} $d_{ct}$ (countertop), $d_{fr}$ (fridge), $d_{si}$ (sink), $d_{c1}$ (cabinet for plate), $d_{c2}$ (cabinet for oil), $d_x$ (to object $x$)
    \item \textit{State booleans:} $\text{Open}(c)$, $\text{Tog}(s)$, $\text{Soaked}$, $\text{DustFree}(c)$, $\text{Clean}(p)$, $\text{OilIn}(c_2)$, $\text{Diff}$
    \item \textit{Counters:} $N_b$ (blenders not on countertop), $N_a$ (apples not in fridge), $N_{cas}$ (casseroles not in fridge), $N_s$ (soaps not next to sink), $N_{cab}$ (cabinets not dusted), $N_r$ (rags not next to sink), $N_{p\_clean}$ (plates not clean), $N_{oil}$ (oil not in cabinet), $N_{p\_store}$ (plates not stored), $N_{close}$ (closed furniture)
\end{itemize}

\resizebox{\columnwidth}{!}{$
\noindent\textit{Open Cabinet and Fridge:}
\begin{aligned}
\{N_{close} > 0, d_c>0\} &\mapsto \{d_c\downarrow\} && \text{; move toward closed furniture} \\
\{N_{close} > 0, d_c = 0\} &\mapsto \{N_{close}\downarrow\}
\end{aligned}
$}

\resizebox{\columnwidth}{!}{$
\noindent\textit{Oil placement:}
\begin{aligned}
\{N_{close} = 0,\neg H_o, d_{oil}>0\} &\mapsto \{d_{oil}\downarrow\} && \text{; approach oil} \\
\{N_{close} = 0,\neg H_o, d_{oil}=0\} &\mapsto \{H_o\} && \text{; pick up oil} \\
\{N_{close} = 0, H_o, N_{oil} > 0, d_{c2}>0\} &\mapsto \{d_{c2}\downarrow\} && \text{; go to cabinet } c_2 \\
\{N_{close} = 0, H_o, N_{oil} > 0, d_{c2} = 0\} &\mapsto \{N_{oil}\downarrow, \neg H_o\} && \text{; place oil in } c_2
\end{aligned}
$}

\resizebox{\columnwidth}{!}{$
\noindent\textit{Plate cleaning:}
\begin{aligned}
\{\neg \text{Tog}(s), d_{si}>0\} &\mapsto \{d_{si}\downarrow\} && \text{; toggle sink on} \\
\{\neg \text{Tog}(s), d_{si}=0\} &\mapsto \{\text{Tog}(s)\} \\
\{\neg H_r, d_{rag}>0\} &\mapsto \{d_{rag}\downarrow\} && \text{; get rag} \\
\{\neg H_r, d_{rag}=0\} &\mapsto \{H_r\} \\
\{H_r, \neg \text{Soaked}, d_{si}>0\} &\mapsto \{d_{si}\downarrow\} && \text{; soak rag} \\
\{H_r, \neg \text{Soaked}, d_{si}=0\} &\mapsto \{\text{Soaked}\} \\
\{H_r, \text{Soaked}, N_{p\_clean}>0, d_{plate}>0\} &\mapsto \{d_{plate}\downarrow\} \\
\{H_r, \text{Soaked}, N_{p\_clean}>0, d_{plate}=0\} &\mapsto \{N_{p\_clean}\downarrow, d_{plate}?\}
\end{aligned}
$}
\resizebox{\columnwidth}{!}{$
\noindent\textit{Plate placement distinct from oil:}
\begin{aligned}
\{N_{close} =0 , N_{oil}=0, N_{p\_clean} =0 ,\neg H_p, N_{p\_store}>0, d_{plate}>0\} &\mapsto \{d_{plate}\downarrow\} \\
\{N_{close} =0 , N_{oil}=0, N_{p\_clean} =0 , \neg H_p, N_{p\_store}>0, d_{plate}=0\} &\mapsto \{H_p\} \\
\{N_{close} =0 , N_{oil}=0, N_{p\_clean} =0 , H_p, N_{p\_store}>0, d_{c1}>0\} &\mapsto \{d_{c1}\downarrow\} \\
\{N_{close} =0 , N_{oil}=0, N_{p\_clean} =0 , H_p, N_{p\_store}>0, d_{c1}=0\} &\mapsto \{N_{p\_store}\downarrow, \neg H_p\}
\end{aligned}
$}

\resizebox{\columnwidth}{!}{$
\noindent\textit{Blender to countertop:}
\begin{aligned}
\{N_b > 0, \neg H_b, d_{blender}>0\} &\mapsto \{d_{blender}\downarrow\} \\
\{N_b > 0, \neg H_b, d_{blender}=0\} &\mapsto \{H_b\} \\
\{H_b, N_b>0, d_{ct}>0\} &\mapsto \{d_{ct}\downarrow\} \\
\{H_b, N_b>0, d_{ct}=0\} &\mapsto \{N_b\downarrow, \neg H_b\}
\end{aligned}
$}

\resizebox{\columnwidth}{!}{$
\noindent\textit{Apple to fridge:}
\begin{aligned}
\{N_{close} =0,\neg H_f, d_{apple}>0\} &\mapsto \{d_{apple}\downarrow\} \\
\{N_{close} =0,\neg H_f, d_{apple}=0\} &\mapsto \{H_f\} \\
\{N_{close} =0,H_f, N_a>0, d_{fr}>0\} &\mapsto \{d_{fr}\downarrow\} \\
\{N_{close} =0,H_f, N_a>0, d_{fr}=0\} &\mapsto \{N_a\downarrow, \neg H_f\}
\end{aligned}
$}

\resizebox{\columnwidth}{!}{$
\noindent\textit{Casserole to fridge:}
\begin{aligned}
\{N_{close} =0 ,\neg H_f, d_{casserole}>0\} &\mapsto \{d_{casserole}\downarrow\} \\
\{N_{close} =0 ,\neg H_f, d_{casserole}=0\} &\mapsto \{H_f\} \\
\{N_{close} =0 ,H_f, N_{cas}>0, d_{fr}>0\} &\mapsto \{d_{fr}\downarrow\} \\
\{N_{close} =0 ,H_f, N_{cas}>0, d_{fr}=0\} &\mapsto \{N_{cas}\downarrow, \neg H_f\}
\end{aligned}
$}

\resizebox{\columnwidth}{!}{$
\noindent\textit{Cabinet dusting:}
\begin{aligned}
\{\neg H_r, d_{rag}>0\} &\mapsto \{d_{rag}\downarrow\} && \text{; get rag} \\
\{\neg H_r, d_{rag}=0\} &\mapsto \{H_r\} \\
\{H_r, N_{cab}>0, d_c>0\} &\mapsto \{d_c\downarrow\} && \text{; clean cabinet} \\
\{H_r, N_{cab}>0, d_c=0\} &\mapsto \{N_{cab}\downarrow, \neg H_r\}
\end{aligned}
$}

\resizebox{\columnwidth}{!}{$
\noindent\textit{Rag near/inside sink:}
\begin{aligned}
\{N_{cab} = 0, N_{p\_clean} = 0,\neg H_r, d_{rag}>0\} &\mapsto \{d_{rag}\downarrow\} \\
\{N_{cab} = 0, N_{p\_clean} = 0,\neg H_r, d_{rag}=0\} &\mapsto \{H_r\} \\
\{N_{cab} = 0, N_{p\_clean} = 0,H_r, N_r>0, d_{si}>0\} &\mapsto \{d_{si}\downarrow\} \\
\{N_{cab} = 0, N_{p\_clean} = 0,H_r, N_r>0, d_{si}=0\} &\mapsto \{N_r\downarrow, \neg H_r\}
\end{aligned}
$}

\resizebox{\columnwidth}{!}{$
\noindent\textit{Soap next to sink:}
\begin{aligned}
\{N_{cab} = 0, N_{p\_clean} = 0,\neg H_s, d_{soap}>0\} &\mapsto \{d_{soap}\downarrow\} \\
\{N_{cab} = 0, N_{p\_clean} = 0,\neg H_s, d_{soap}=0\} &\mapsto \{H_s\} \\
\{N_{cab} = 0, N_{p\_clean} = 0,H_s, N_s>0, d_{si}>0\} &\mapsto \{d_{si}\downarrow\} \\
\{N_{cab} = 0, N_{p\_clean} = 0,H_s, N_s>0, d_{si}=0\} &\mapsto \{N_s\downarrow, \neg H_s\}
\end{aligned}
$}

\subsection{Task 4: Organizing File Cabinet}

\textbf{Features:}
\begin{itemize}[noitemsep]
    \item $H_m$: holding a marker
    \item $H_f$: holding a folder
    \item $H_d$: holding a document
    \item $p_m$: distance to the nearest marker
    \item $p_f$: distance to the nearest folder
    \item $p_d$: distance to the nearest document
    \item $p_t$: distance to a table
    \item $p_c$: distance to a valid cabinet target
    \item $N_m$: number of markers not on a table
    \item $N_f$: number of folders not in a cabinet
    \item $N_d$: number of documents not in a cabinet
    \item $N_o$: number of closed cabinets
\end{itemize}

\resizebox{\columnwidth}{!}{$
\noindent\textit{Opening Cabinet:}
\begin{aligned}
\{N_o > 0, p_c > 0\} &\mapsto \{p_c \downarrow\} && \text{; move toward a closed cabinet} \\
\{N_o > 0, p_c = 0\} &\mapsto \{N_o \downarrow\} && \text{; open the cabinet when reachable}
\end{aligned}
$}
\resizebox{\columnwidth}{!}{$
\noindent\textit{Marker task:}
\begin{aligned}
\{N_m > 0, \neg H_m, p_m > 0\} &\mapsto \{p_m \downarrow, p_t ?\} && \text{; move toward the nearest marker} \\
\{N_m > 0, \neg H_m, p_m = 0\} &\mapsto \{H_m, p_t ?\} && \text{; pick up the marker when reachable} \\
\{N_m > 0, H_m, p_t > 0\} &\mapsto \{p_t \downarrow\} && \text{; move toward a table} \\
\{N_m > 0, H_m, p_t = 0\} &\mapsto \{N_m \downarrow, \neg H_m, p_m ?\} && \text{; put marker on table}
\end{aligned}
$}
\resizebox{\columnwidth}{!}{$
\noindent\textit{Folder task:}
\begin{aligned}
\{N_f > 0, N_o = 0, \neg H_f, p_f > 0\} &\mapsto \{p_f \downarrow, p_c ?\} \\
\{N_f > 0, N_o = 0, \neg H_f, p_f = 0\} &\mapsto \{H_f, p_c ?\} \\
\{N_f > 0, N_o = 0, H_f, p_c > 0\} &\mapsto \{p_c \downarrow\} \\
\{N_f > 0, N_o = 0, H_f, p_c = 0\} &\mapsto \{N_f \downarrow, \neg H_f, p_f ?\}
\end{aligned}
$}
\resizebox{\columnwidth}{!}{$
\noindent\textit{Document task:}
\begin{aligned}
\{N_d > 0, N_o = 0, \neg H_d, p_d > 0\} &\mapsto \{p_d \downarrow, p_c ?\} \\
\{N_d > 0, N_o = 0, \neg H_d, p_d = 0\} &\mapsto \{H_d, p_c ?\} \\
\{N_d > 0, N_o = 0, H_d, p_c > 0\} &\mapsto \{p_c \downarrow\} \\
\{N_d > 0, N_o = 0, H_d, p_c = 0\} &\mapsto \{N_d \downarrow, \neg H_d, p_d ?\}
\end{aligned}
$}

\subsection{Task 5: Thawing Frozen Food}

\textbf{Features:}
\begin{itemize}[noitemsep]
    \item $H_{fish}$: holding a fish
    \item $H_{olive}$: holding an olive
    \item $H_{date}$: holding a date
    \item $p_{fish}$: distance to the nearest fish
    \item $p_{olive}$: distance to the nearest olive
    \item $p_{date}$: distance to the nearest date
    \item $p_{sink}$: distance to a sink
    \item $N_{fish}$: number of fish not next to a sink
    \item $N_{olive}$: number of olives not next to a sink
    \item $N_{date}$: number of dates not next to a fish
\end{itemize}

\resizebox{\columnwidth}{!}{$
\noindent\textit{Fish Task:}
\begin{aligned}
\{\neg H_{fish}, p_{fish} > 0\} &\mapsto \{p_{fish} \downarrow\} \\
\{\neg H_{fish}, p_{fish} = 0\} &\mapsto \{H_{fish}\} \\
\{H_{fish}, p_{sink} > 0\} &\mapsto \{p_{sink} \downarrow\} \\
\{H_{fish}, p_{sink} = 0\} &\mapsto \{N_{fish} \downarrow, \neg H_{fish}, p_{fish} ?\}
\end{aligned}
$}

\resizebox{\columnwidth}{!}{$
\noindent\textit{Olive Task:}
\begin{aligned}
\{\neg H_{olive}, p_{olive} > 0\} &\mapsto \{p_{olive} \downarrow\} \\
\{\neg H_{olive}, p_{olive} = 0\} &\mapsto \{H_{olive}\} \\
\{H_{olive}, p_{sink} > 0\} &\mapsto \{p_{sink} \downarrow\} \\
\{H_{olive}, p_{sink} = 0\} &\mapsto \{N_{olive} \downarrow, \neg H_{olive}, p_{olive} ?\}
\end{aligned}
$}

\resizebox{\columnwidth}{!}{$
\noindent\textit{Date Task:}
\begin{aligned}
\{\neg H_{date}, N_{\text{fish olive}}= 0, p_{date} > 0\} &\mapsto \{p_{date} \downarrow\} \\
\{\neg H_{date}, N_{\text{fish olive}}= 0,p_{date} = 0\} &\mapsto \{H_{date}\} \\
\{H_{date}, N_{\text{fish olive}}= 0,p_{fish} > 0\} &\mapsto \{p_{fish} \downarrow\} \\
\{H_{date}, N_{\text{fish olive}}= 0, p_{fish} = 0\} &\mapsto \{N_{date} \downarrow, \neg H_{date}, p_{date} ?\}
\end{aligned}
$}

\subsection{Task 6: Making Tea}

\textbf{Features:}
\begin{itemize}[noitemsep]
    \item $H_l$: holding a lemon
    \item $H_{knife}$: holding a knife
    \item $H_t$: holding a teapot
    \item $H_{tb}$: holding a tea bag
    \item $p_l$: distance to the nearest lemon
    \item $p_{knife}$: distance to the knife
    \item $p_t$: distance to the nearest teapot
    \item $p_{tb}$: distance to the nearest tea bag
    \item $p_s$: distance to the stove
    \item $p_{sink}$: distance to the sink
    \item $\text{is\_sliced}(\text{lemon})$: whether the lemon is sliced
    \item $\text{is\_toggled}(\text{stove})$: whether the stove is toggled on
    \item $\text{is\_toggled}(\text{sink})$: whether the sink is toggled on
    \item $\text{is\_soaked}(\text{tea\_bag})$: whether the tea bag is soaked
    \item $\text{atsamelocation}(\text{tea\_bag}, \text{teapot})$: tea bag at same location as teapot
    \item $\text{onTop}(\text{teapot}, \text{stove})$: teapot on stove
\end{itemize}

\resizebox{\columnwidth}{!}{$
\noindent\textit{Toggle stove:}
\begin{aligned}
\{\neg \text{is\_toggled}(\text{stove}), p_s > 0\} &\mapsto \{p_s \downarrow\} \\
\{\neg \text{is\_toggled}(\text{stove}), p_s = 0\} &\mapsto \{\text{is\_toggled}(\text{stove})\}
\end{aligned}
$}
\resizebox{\columnwidth}{!}{$
\noindent\textit{Slice lemon:}
\begin{aligned}
\{\neg H_{knife}, p_{knife} > 0, \neg \text{is\_sliced}(\text{lemon})\} &\mapsto \{p_{knife} \downarrow\} \\
\{\neg H_{knife}, p_{knife} = 0, \neg \text{is\_sliced}(\text{lemon})\} &\mapsto \{H_{knife}\} \\
\{H_{knife}, \neg \text{is\_sliced}(\text{lemon}), p_{l} > 0\} &\mapsto \{p_{l} \downarrow\} \\
\{H_{knife}, \neg \text{is\_sliced}(\text{lemon}), p_{l} = 0\} &\mapsto \{\text{is\_sliced}(\text{lemon})\} \\
\{H_{knife}, \text{is\_sliced}(\text{lemon})\} &\mapsto \{\neg H_{knife}, p_{knife} ?\}
\end{aligned}
$}
\resizebox{\columnwidth}{!}{$
\noindent\textit{Handle teapot:}
\begin{aligned}
\{\neg \text{atsameloc}(\text{tb}, \text{tp}), \neg \text{onTop}(\text{tp}, \text{st}), \neg H_t, p_t > 0\} &\mapsto \{p_t \downarrow\} \\
\{\neg \text{atsameloc}(\text{tb}, \text{tp}), \neg \text{onTop}(\text{tp}, \text{st}), \neg H_t, p_t = 0\} &\mapsto \{H_t\} \\
\{\neg \text{atsameloc}(\text{tb}, \text{tp}), \neg \text{onTop}(\text{tp}, \text{st}), H_t, p_s > 0\} &\mapsto \{p_s \downarrow\} \\
\{\neg \text{atsameloc}(\text{tb}, \text{tp}), \neg \text{onTop}(\text{tp}, \text{st}), H_t, p_s = 0\} &\mapsto \{\text{onTop}(\text{tp}, \text{st}), \neg H_t, p_t ?\} \\
\{\neg \text{atsameloc}(\text{tb}, \text{tp}), \text{onTop}(\text{tp}, \text{st}), \neg H_{tb}, p_{tb} > 0\} &\mapsto \{p_{tb} \downarrow\} \\
\{\neg \text{atsameloc}(\text{tb}, \text{tp}), \text{onTop}(\text{tp}, \text{st}), \neg H_{tb}, p_{tb} = 0\} &\mapsto \{H_{tb}\} \\
\{\neg \text{atsameloc}(\text{tb}, \text{tp}), \text{onTop}(\text{tp}, \text{st}), H_{tb}, p_t > 0\} &\mapsto \{p_t \downarrow\} \\
\{H_{tb}, \neg \text{atsameloc}(\text{tb}, \text{tp}), \text{onTop}(\text{tp}, \text{st}), p_t = 0\} &\mapsto \{\text{atsameloc}(\text{tb}, \text{tp}), \neg H_{tb}, p_{tb} ?\}
\end{aligned}
$}

\subsection{Task 7: Opening Packages}

\textbf{Features:}
\begin{itemize}[noitemsep]
    \item $p$: distance to the nearest package
    \item $N$: number of unopened packages
\end{itemize}

\resizebox{\columnwidth}{!}{$
\begin{aligned}
\{N > 0, p > 0\} &\mapsto \{p \downarrow\} && \text{; move toward the nearest unopened package} \\
\{N > 0, p = 0\} &\mapsto \{N \downarrow\} && \text{; open the package when reachable}
\end{aligned}
$}

\subsection{Task 8: Boxing Books Up for Storage}

\textbf{Features:}
\begin{itemize}[noitemsep]
    \item $H$: holding an unboxed book
    \item $p$: distance to the nearest unboxed book
    \item $b$: distance to a valid box target
    \item $N$: number of unboxed books
\end{itemize}

\resizebox{\columnwidth}{!}{$
\begin{aligned}
\{\neg H, p > 0\} &\mapsto \{p \downarrow, b ?\} && \text{; move toward the nearest unboxed book} \\
\{\neg H, p = 0\} &\mapsto \{H\} && \text{; pick up the book when reachable} \\
\{H, N > 0, b > 0\} &\mapsto \{b \downarrow\} && \text{; move toward a chosen box spot} \\
\{H, N > 0, b = 0\} &\mapsto \{\neg H, N \downarrow, p ?\} && \text{; place the book into the box}
\end{aligned}
$}

\subsection{Task 9: Collect Misplaced Items}

\textbf{Features:}
\begin{itemize}[noitemsep]
    \item $H_s$: holding a gym shoe
    \item $H_n$: holding a necklace
    \item $H_b$: holding a notebook
    \item $H_k$: holding a sock
    \item $p_s$: distance to the nearest gym shoe
    \item $p_n$: distance to the nearest necklace
    \item $p_b$: distance to the nearest notebook
    \item $p_k$: distance to the nearest sock
    \item $p_t$: distance to a table
    \item $N_s$: number of gym shoes not on a table
    \item $N_n$: number of necklaces not on a table
    \item $N_b$: number of notebooks not on a table
    \item $N_k$: number of socks not on a table
\end{itemize}

\resizebox{\columnwidth}{!}{$
\noindent\textit{Gym shoe task:}
\begin{aligned}
\{\neg H_s, N_s > 0, p_s > 0\} &\mapsto \{p_s \downarrow, p_t ?\} \\
\{\neg H_s, N_s > 0, p_s = 0\} &\mapsto \{H_s, p_t ?\} \\
\{H_s, N_s > 0, p_t > 0\} &\mapsto \{p_t \downarrow\} \\
\{H_s, N_s > 0, p_t = 0\} &\mapsto \{N_s \downarrow, \neg H_s, p_s ?\}
\end{aligned}
$}
\resizebox{\columnwidth}{!}{$
\noindent\textit{Necklace task:}
\begin{aligned}
\{\neg H_n, N_n > 0, p_n > 0\} &\mapsto \{p_n \downarrow, p_t ?\} \\
\{\neg H_n, N_n > 0, p_n = 0\} &\mapsto \{H_n, p_t ?\} \\
\{H_n, N_n > 0, p_t > 0\} &\mapsto \{p_t \downarrow\} \\
\{H_n, N_n > 0, p_t = 0\} &\mapsto \{N_n \downarrow, \neg H_n, p_n ?\}
\end{aligned}
$}
\resizebox{\columnwidth}{!}{$
\noindent\textit{Notebook task:}
\begin{aligned}
\{\neg H_b, N_b > 0, p_b > 0\} &\mapsto \{p_b \downarrow, p_t ?\} \\
\{\neg H_b, N_b > 0, p_b = 0\} &\mapsto \{H_b, p_t ?\} \\
\{H_b, N_b > 0, p_t > 0\} &\mapsto \{p_t \downarrow\} \\
\{H_b, N_b > 0, p_t = 0\} &\mapsto \{N_b \downarrow, \neg H_b, p_b ?\}
\end{aligned}
$}
\resizebox{\columnwidth}{!}{$
\noindent\textit{Sock task:}
\begin{aligned}
\{\neg H_k, N_k > 0, p_k > 0\} &\mapsto \{p_k \downarrow, p_t ?\} \\
\{\neg H_k, N_k > 0, p_k = 0\} &\mapsto \{H_k, p_t ?\} \\
\{H_k, N_k > 0, p_t > 0\} &\mapsto \{p_t \downarrow\} \\
\{H_k, N_k > 0, p_t = 0\} &\mapsto \{N_k \downarrow, \neg H_k, p_k ?\}
\end{aligned}
$}

\subsection{Task 10: Putting Away Dishes After Cleaning}

\textbf{Features:}
\begin{itemize}[noitemsep]
    \item $H$: holding a plate
    \item $p$: distance to the nearest plate
    \item $b$: distance to a valid cabinet target
    \item $N$: number of unboxed plates
    \item $\text{is\_opened}$: whether the cabinet is opened
\end{itemize}

\resizebox{\columnwidth}{!}{$
\begin{aligned}
\{\neg \text{is\_opened}, b > 0\} &\mapsto \{b \downarrow\} && \text{; move toward the cabinet} \\
\{\neg \text{is\_opened}, b = 0\} &\mapsto \{\text{is\_opened}\} && \text{; open the cabinet when reachable} \\
\{\neg H, N > 0, p > 0, \text{is\_opened}\} &\mapsto \{p \downarrow, b ?\} \\
\{\neg H, N > 0, p = 0, \text{is\_opened}\} &\mapsto \{H, b ?\} \\
\{H, N > 0, b > 0, \text{is\_opened}\} &\mapsto \{b \downarrow\} \\
\{H, N > 0, b = 0, \text{is\_opened}\} &\mapsto \{\neg H, N \downarrow, p ?\}
\end{aligned}
$}

\subsection{Instruction Generation Statistics} We used a template-based generator to translate the symbolic rule sets into natural language instructions, ensuring alignment with the VLA's training data format. Table \ref{tab:instruction_stats} presents the summary statistics for the generated instructions across the three granularity levels.

\begin{table*}[t] 
\centering 
\caption{Statistics of generated natural language instructions across granularity levels.} 
\label{tab:instruction_stats} 
\resizebox{\textwidth}{!}{
\begin{tabular}{l|c|c|c|c|c|c|c|c}
\toprule 
\textbf{Granularity} & \textbf{Count} & \textbf{Avg Length} & \textbf{Std Length} & \textbf{Min} & \textbf{Max} & \textbf{Unique Tokens} & \textbf{TTR} & \textbf{Avg Sent. Length} \\ 
 & & \textbf{(words)} & & & & & \textbf{(Type-Token Ratio)} & \\ 
\midrule
\textbf{Fine (F)} & 1636 & 41.12 & 11.36 & 19 & 87 & 140 & 0.0021 & 41.12 \\ 
\textbf{Medium (M)} & 859 & 38.62 & 12.76 & 22 & 81 & 108 & 0.0033 & 38.62 \\ 
\textbf{Coarse (C)} & 587 & 33.60 & 13.39 & 22 & 95 & 94 & 0.0048 & 33.60 \\ 
\bottomrule 
\end{tabular}
} 
\end{table*}

\section{Implementation Details}
\label{app_sec:implementation_details}

\subsection{Model Architecture}

Our policy is built upon the OpenVLA architecture \citep{DBLP:conf/corl/KimPKXB0RFSVKBT24}. Specifically, we utilize the Prismatic-7B backbone \citep{DBLP:conf/icml/Karamcheti0BLKS24}, which integrates a frozen Llama-2-7B language model with fused visual features. The visual encoder consists of a dual-stream setup combining SigLIP (for semantic understanding) and DINOv2 (for fine-grained spatial geometry) \citep{DBLP:conf/iccv/ZhaiM0B23,DBLP:journals/tmlr/OquabDMVSKFHMEA24}.

\subsection{Training Configuration}

We fine-tune the model using Low-Rank Adaptation (LoRA) on the language model backbone, keeping the visual encoders frozen. All training setting variants (Pure, Centroid, Axial) share the same underlying training hyperparameters to ensure fair comparison.

Hardware and Duration: Training was conducted on a single node equipped with 8$\times$ NVIDIA A100 GPUs. A standard training run for 100,000 steps took approximately 23 hours to complete.

Table \ref{tab:hyperparameters} summarizes the core hyperparameters used across our experiments.

\begin{table}[H]
\centering
\caption{Hyperparameters for OpenVLA fine-tuning on \textsc{Mini-BEHAVIOR-Gran}.}
\label{tab:hyperparameters}
\resizebox{\columnwidth}{!}{
\begin{tabular}{l|c}
\toprule
\textbf{Hyperparameter} & \textbf{Value} \\
\midrule
Base Model & Prismatic-7B (Llama-2) \\
Visual Encoders & SigLIP + DINOv2 \\
Parameter Efficient Tuning & LoRA (Rank $r=32$) \\
Compute Hardware & 8$\times$ NVIDIA A100 \\
Training Duration & $\sim$23 hours \\
Optimization Steps & 100,000 \\
Steps Before Decay & 80,000 \\
Learning Rate & $2.5 \times 10^{-4}$ \\
Batch Size (per GPU) & $7 - 8$ \\
\bottomrule
\end{tabular} 
}
\end{table}

We also train a lightweight VLM predictor to quantify instruction predictability via perplexity. The model is fine-tuned using Low-Rank Adaptation (LoRA) on the Qwen3-VL-4B-Instruct backbone, with vision encoders frozen to preserve visual grounding capabilities established during pretraining. All models are trained on \textsc{Mini-BEHAVIOR-Gran} instruction and image training data pairs, with training and evaluation splits maintained consistently across experiments. Hardware and Duration: Training was conducted on a single node equipped with 4$\times$ NVIDIA RTX 5090 GPUs. Each training run for 2 epochs. Table \ref{tab:qwen_hyperparameters} summarizes the core hyperparameters used for VLM predictor training.

\begin{table}[H]
\centering
\caption{Hyperparameters for Qwen3-VL-4B-Instruct fine-tuning on \textsc{Mini-BEHAVIOR-Gran} (High-domain).}
\label{tab:qwen_hyperparameters}
\resizebox{\columnwidth}{!}{
\begin{tabular}{l|c}
\toprule
\textbf{Hyperparameter} & \textbf{Value} \\
\midrule
Base Model & Qwen3-VL-4B-Instruct \\
Visual Encoders & Frozen (Qwen3-VL native) \\
Parameter Efficient Tuning & LoRA (Rank $r=64$, $\alpha=128$) \\
LoRA Target Modules & All linear layers \\
Compute Hardware & 4$\times$ NVIDIA RTX 5090 \\
Training Epochs & 2 \\
Learning Rate & $1.0 \times 10^{-5}$ \\
Learning Rate Scheduler & Cosine with 10\% warmup \\
Batch Size (per GPU) & 32 \\
Seed & [42,50,60] \\
Gradient Accumulation Steps & 1 \\
DeepSpeed Strategy & Zero-2 \\
\bottomrule
\end{tabular} 
}
\end{table}

\section{Additional Experimental Results}
\label{app_sec:additional_experimental_results}

This section provides additional detailed per-task performance breakdowns, and failure mode distributions.

\subsection{Detailed Task Performance} 

\begin{figure*}[t]
    \centering
    \includegraphics[width=\linewidth]{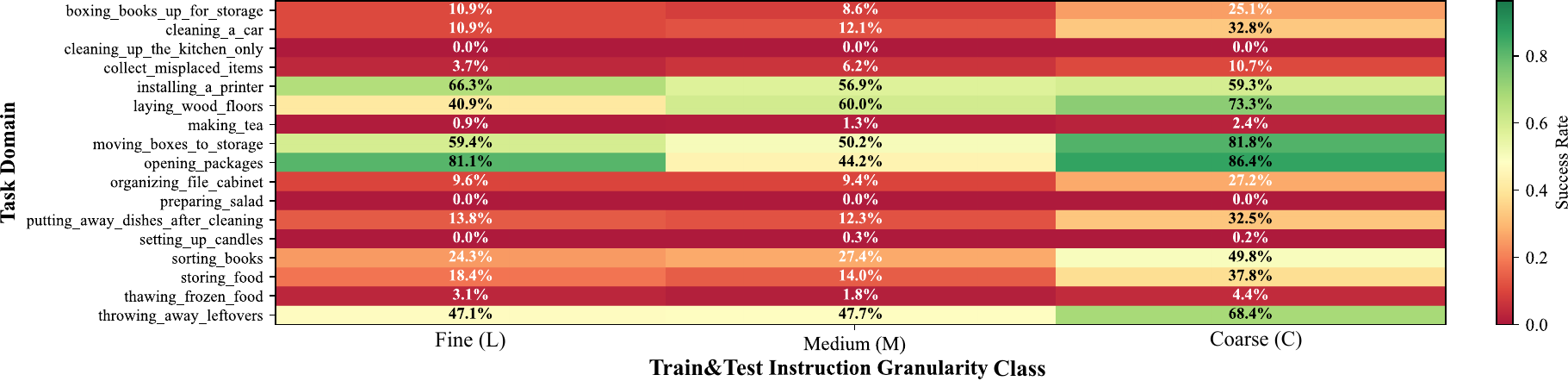}
    \caption{Detailed per-task success rates.}
    \label{fig:per_task_success_rate}
\end{figure*}

Figure \ref{fig:per_task_success_rate} details the success rates across all 20 tasks in the benchmark. We see some tasks are too difficult to solve and thus low across all instruction granularity. And then we also see the U shape trends in most tasks, showing the same phenomenon we discussed in \textbf{(Q1)} that both very fine and very coarse instructions are easier for VLA to follow.

\subsection{Detailed Attention Analysis Table}
\label{app_sec:full_attention_analysis_table}

We quantify the internal information-gathering behavior of Vision-Language-Action (VLA) models by analyzing the \textbf{attention weight distribution of the action tokens}. In these architectures, the action tokens serve as the terminal query point; their attention allocation reveals which input modalities (vision vs. language) the model prioritizes at different stages of its internal processing hierarchy.

\subsubsection{Input Sequence Formalization}
The input sequence $\mathcal{S}$ is composed of visual patches, linguistic tokens, and action tokens. Given a single sample, the sequence layout is defined as:
\begin{equation}
    \mathcal{S} = [ \texttt{BOS}, \underbrace{v_1, \dots, v_P}_{\text{vision}}, \underbrace{t_1, \dots, t_T}_{\text{text}}, \underbrace{a_1, \dots, a_A}_{\text{action}}, \texttt{EOS} ]
\end{equation}
where $P$ is the number of visual patch tokens, $T$ is the number of text tokens (comprising both prompt prefix and task instruction), and $A$ is the number of action tokens (where $A = d_{\text{action}} \times \text{chunk size}$). 

For analysis, we exclude the $[\texttt{BOS}]$ and $[\texttt{EOS}]$ boundary tokens. For each layer $\ell \in \{1, \dots, L\}$, we extract the trimmed attention tensor:
\begin{equation}
    \mathbf{A}^{(\ell)} \in \mathrm{R}^{H \times (P+T+A) \times (P+T+A)}
\end{equation}

\subsubsection{Partitioning of Semantic Regions}
To distinguish between general linguistic context and specific task instructions, we partition the key dimension (columns) of $\mathbf{A}^{(\ell)}$ into four non-overlapping semantic regions:

\textit{Note: For models utilizing compressed instruction patches, the compressed embeddings are mapped to the Dynamic Text category to ensure functional consistency across architectures.}

\subsubsection{Quantitative Measurement}
For each layer $\ell$ and head $h$, we isolate the \textbf{action-to-key} sub-matrix $\mathbf{A}^{(\ell)}_{h,\text{action} \to \text{key}} \in \mathrm{R}^{A \times (P+T+A)}$. The mean attention mass allocated to a specific region $R = [r_{\text{start}}, r_{\text{end}})$ is computed by summing weights across the region and averaging over the $A$ action tokens:
\begin{equation}
    \text{attn}_h(R) = \frac{1}{A} \sum_{i=1}^{A} \sum_{j=r_{\text{start}}}^{r_{\text{end}}-1} \mathbf{A}^{(\ell)}_{h,\text{action} \to \text{key}}[i, j]
\end{equation}

\subsubsection{Aggregation and Normalization}
To derive a layer-wise modality reliance score, we perform the following:
\begin{enumerate}
    \item \textbf{Head Averaging}: $\text{attn}(R) = \frac{1}{H} \sum_{h=1}^{H} \text{attn}_h(R)$.
    \item \textbf{Sample Mean}: Values are averaged across all samples within a specific granularity-model pair.
    \item \textbf{Simplex Normalization}: The final regional shares $\hat{a}_R$ are normalized such that $\sum_{R} \hat{a}_R = 1$.
\end{enumerate}

\subsubsection{Statistical Trend Hypothesis Testing}
We define three binary indicators to test the \textbf{Shallow Grounding Hypothesis} layer-wise:
\begin{itemize}
    \item \textbf{T1 (Lang $\uparrow$)}: $\hat{a}_{\text{text}}(\text{Medium}) > \hat{a}_{\text{text}}(\text{Fine})$
    \item \textbf{T2 (Lang $\downarrow$)}: $\hat{a}_{\text{text}}(\text{Medium}) > \hat{a}_{\text{text}}(\text{Coarse})$
    \item \textbf{T3 (Vis $\uparrow$)}: $\hat{a}_{\text{vis}}(\text{Coarse}) > \hat{a}_{\text{vis}}(\text{Medium})$
\end{itemize}
We assess the significance of these trends across $N=32$ layers using a \textbf{one-sided Binomial test} ($H_0: p=0.5$). Significance is reported as $^{*}p < 0.05$ and $^{**}p < 0.01$. Confidence intervals for proportions are calculated using the \textbf{Clopper-Pearson} method. The full statistical breakdown is presented in \Cref{tab:full_stats_appendix}.

\begin{table*}[h]
\centering
\small
\begin{tabular}{ll cccc c}
\toprule
\textbf{Model} & \textbf{Trend} & \textbf{k/N} & \textbf{Prop.} & \textbf{95\% CI} & \textbf{$H_0$} & \textbf{$p$-value} \\
\midrule
Autoregressive     & T1 & 25/32 & 78.1\% & [0.600, 0.907] & 50.0\% & 0.0011 \\
                   & T2 & 4/32  & 12.5\% & [0.035, 0.290] & 50.0\% & 1.0000 \\
                   & T3 & 21/32 & 65.6\% & [0.468, 0.814] & 50.0\% & 0.1102 \\
\midrule
Parallel Decoding  & T1 & 5/32  & 15.6\% & [0.053, 0.328] & 50.0\% & 1.0000 \\
                   & T2 & 9/32  & 28.1\% & [0.137, 0.467] & 50.0\% & 0.9965 \\
                   & T3 & 5/32  & 15.6\% & [0.053, 0.328] & 50.0\% & 1.0000 \\
\midrule
Discrete Diffusion & T1 & 17/32 & 53.1\% & [0.347, 0.709] & 50.0\% & 0.4300 \\
                   & T2 & 23/32 & 71.9\% & [0.533, 0.863] & 50.0\% & 0.0100 \\
                   & T3 & 24/32 & 75.0\% & [0.566, 0.885] & 50.0\% & 0.0035 \\
\bottomrule
\end{tabular}
\caption{Detailed statistical analysis of layer-wise attention trends across architectures. P-values are calculated using a one-sided Binomial test against a null hypothesis ($H_0$) of 0.5. Confidence intervals (CI) are calculated at the 95\% level.}
\label{tab:full_stats_appendix}
\end{table*}

\subsection{Failure Analysis} 
\begin{figure}[H]
    \centering
    \includegraphics[width=\linewidth]{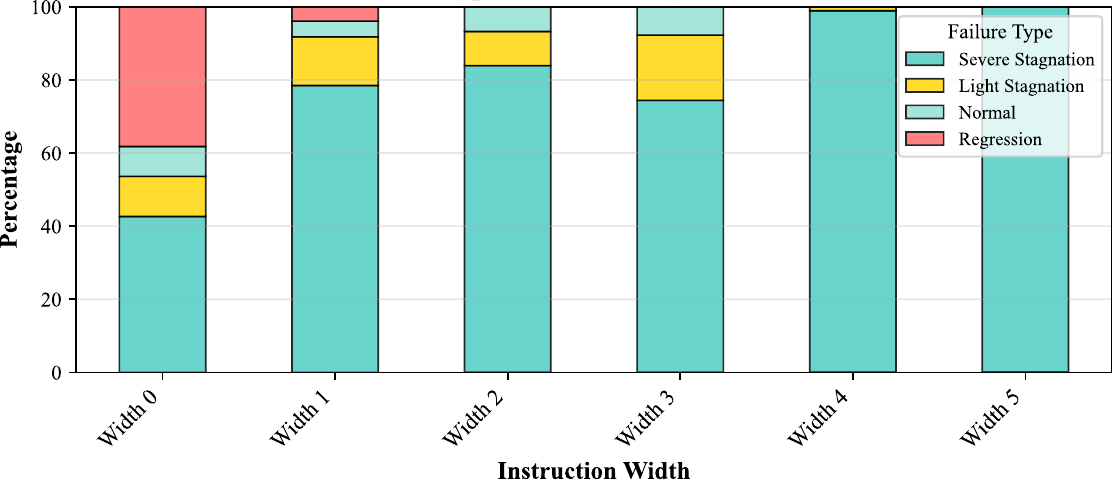}
    \caption{Failure type distribution across instruction widths.}
    \label{fig:failure_type_dist_by_width}
\end{figure}

\begin{figure}[H]
    \centering
    \includegraphics[width=\linewidth, keepaspectratio]{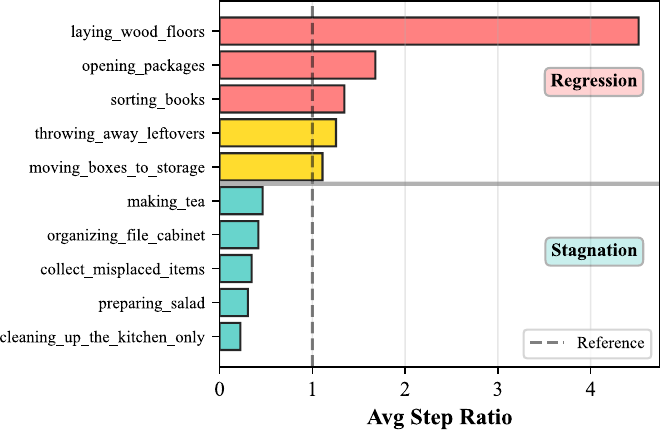}
    \captionof{figure}{Low-width but long task horizon (e.g., laying woods) tends to cause regression, while high-width stagnates.}
    \label{fig:q4_failure_modes}
\end{figure}

Figure \ref{fig:failure_type_dist_by_width} presents the distribution of failure types across varying instruction widths. It confirms that low width encounters regression failures more frequently, while high width predominantly leads to stagnation failures. This aligns with our earlier observations in \Cref{fig:q4_failure_modes} regarding the distinct failure modes associated with different instruction granularities. Agent failure distributions bifurcate strictly with width. Low-width, long-horizon tasks predominantly suffer from \emph{regression}: the agent violates previously satisfied fluents and cycles back to prior states, reflecting subgoal serialization failures. High-width tasks ($w \ge 3$), by contrast, exhibit \emph{stagnation}: the agent freezes, unable to resolve concurrent feature dependencies ($O(|\Phi|^w)$) required to initiate a valid state transition. This dichotomy mirrors the width-based complexity landscape.

\subsection{More Plots for Width vs. Other Linguistic Heuristics as Metric for Language Grounding Complexity}

\begin{figure}[H]
    \centering
    \includegraphics[width=\linewidth]{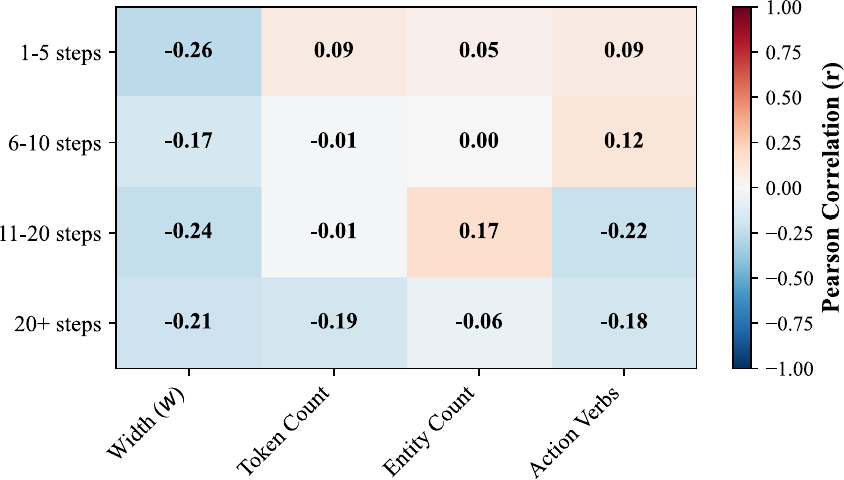}
    \caption{Correlation heatmap between instruction width and other linguistic heuristics.}
    \label{fig:correlation_heatmap}
\end{figure}

As shown in the correlation heatmap \Cref{fig:correlation_heatmap}, $w$ is the only metric that maintains a stable negative correlation with Success Rate across all horizons. In contrast, surface metrics like Token Count and Entity Count exhibit near-zero or even positive correlations in shorter horizons (e.g., $r=0.09$ for 1-5 steps). This confirms the Complexity Paradox: while longer, more detailed instructions (higher token/entity counts) could actually provide helpful guidance that improves SR, increasing the latent planning width ($w$) invariably degrades performance.

\begin{figure}[H]
    \centering
    \includegraphics[width=\linewidth]{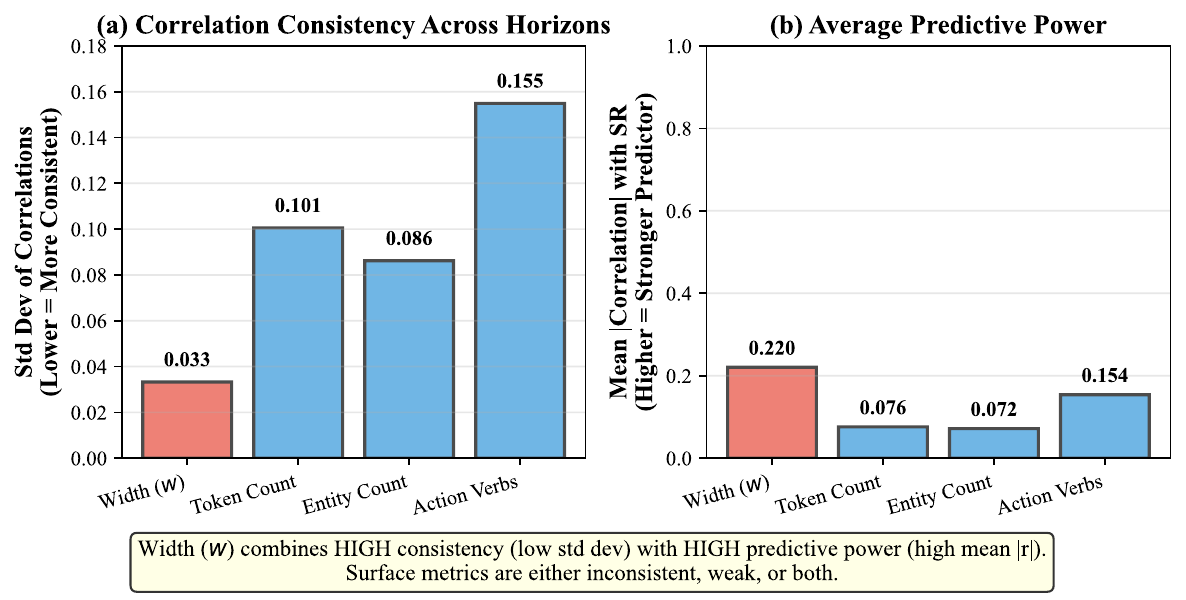}
    \caption{Consistency comparison between instruction width and other linguistic heuristics.}
    \label{fig:consistency_comparison}
\end{figure}

Similarly, the consistency analysis in \Cref{fig:consistency_comparison} shows that $w$ possesses the lowest standard deviation ($\sigma=0.033$), indicating its predictive reliability is invariant to task horizons. Surface metrics, particularly Action Verbs ($\sigma=0.155$), are highly volatile. It also shows that $w$ provides the highest mean predictive power ($|r|=0.220$), more than doubling the predictive strength of Token Count ($0.076$) or Entity Count ($0.072$).

\subsection{Sample Generated Instructions by Granularity} \label{app_sec:sample_instructions}

To illustrate the qualitative differences between granularity levels, we provide five random samples for each category below. Note how Low (L) granularity explicitly specifies navigation and preconditions (e.g., move toward,'' not yet at''), while High (H) granularity focuses almost exclusively on the final state change (e.g., ``reduce the count of...''), shown in Table~\ref{tab:sample_instructions}.

\begin{table*}[ht]
\centering
\begin{minipage}{\textwidth}

\noindent\textbf{Fine Granularity (F) Samples}
\begin{tcolorbox}[colback=green!5, colframe=green!40!black]
\small
\begin{enumerate}
    \item At step 11: When the count of uncleaned plate is above 0, not yet at sink 0 and sink is off, please try to move toward sink 0.
    \item At step 8: When not yet at soap 0, not holding soap 0, the count of unwiped car is 0 and the count of rag not inside bucket is 0, please try to move toward soap 0.
    \item At step 27: When the count of chip and oatmeal sugar and vegetable oil not inside cabinet is above 0, you are at oatmeal 1 and not holding oatmeal 1, please pick up oatmeal 1.
    \item At step 15: When the count of plate not inside cabinet is above 0, you are at plate 3, the count of closed cabinet is 0 and not holding plate 3, please pick up plate 3.
    \item At step 13: When the count of fish and olive not near sink is above 0, not yet at olive 0 and not holding olive 0, please try to move toward olive 0.
\end{enumerate}
\end{tcolorbox}

\noindent\textbf{Medium Granularity (M) Samples}
\begin{tcolorbox}[colback=yellow!5, colframe=yellow!40!black]
\small
\begin{enumerate}
    \item At step 2: When the count of plate not inside cabinet is above 0, the count of closed cabinet is 0 and not holding plate 2, please pick up plate 2.
    \item At step 6: When the count of uncleaned plate is above 0, the count of dry rag is above 0 and holding rag 0, please reduce the count of dry rag.
    \item At step 9: When the count of plate not inside cabinet 0 is above 0, the count of closed cabinet is 0, the count of vegetable oil not inside cabinet 1 is 0, the count of uncleaned plate is 0 and not holding plate 0, please pick up plate 0.
    \item At step 13: When the count of book not inside box is above 0 and not holding book 2, please pick up book 2.
    \item At step 9: When the count of chip and oatmeal sugar and vegetable oil not inside cabinet is above 0, holding chip 1 and the count of closed cabinet is 0, please not holding chip 1, then reduce the count of chip and oatmeal sugar and vegetable oil not inside cabinet.
\end{enumerate}
\end{tcolorbox}

\noindent\textbf{Coarse Granularity (C) Samples}
\begin{tcolorbox}[colback=blue!5, colframe=blue!40!black]
\small
\begin{enumerate}
    \item At step 2: When the count of hamburger not inside ashcan is above 0, please reduce the count of hamburger not inside ashcan.
    \item At step 5: When the count of plate not inside cabinet is above 0 and the count of closed cabinet is 0, please reduce the count of plate not inside cabinet.
    \item At step 3: When the count of teapot not on stove is above 0, please reduce the count of teapot not on stove.
    \item At step 9: When the count of plate not inside cabinet is above 0 and the count of closed cabinet is 0, please reduce the count of plate not inside cabinet.
    \item At step 8: When the count of chip and oatmeal sugar and vegetable oil not inside cabinet is above 0 and the count of closed cabinet is 0, please reduce the count of chip and oatmeal sugar and vegetable oil not inside cabinet.
\end{enumerate}
\end{tcolorbox}

\end{minipage}
\caption{Sample generated instructions across granularity levels.}
\label{tab:sample_instructions}
\end{table*}

\end{document}